\documentclass[10pt,twocolumn,letterpaper]{article}

\usepackage[pagenumbers]{cvpr} 

\usepackage{graphicx}
\usepackage{amsmath}
\usepackage{amssymb}
\usepackage{booktabs}

\usepackage{pifont}
\newcommand{\cmark}{\ding{51}}%
\newcommand{\xmark}{\ding{55}}%
\usepackage{algorithm}
\usepackage{algorithmic}
\usepackage{bbm}
\usepackage[utf8]{inputenc} 
\usepackage[T1]{fontenc}    
\usepackage{url}            
\usepackage{booktabs}       
\usepackage{amsfonts}       
\usepackage{nicefrac}       
\usepackage{microtype}      
\usepackage{xcolor}         
\usepackage{graphicx}
\usepackage{kotex}
\usepackage{multirow}
\usepackage{amsmath}
\usepackage{enumitem,lipsum}
\usepackage{newfloat}
\usepackage{listings}

\usepackage[pagebackref,breaklinks,colorlinks]{hyperref}

\usepackage{marvosym}

\usepackage[capitalize]{cleveref}
\crefname{section}{Sec.}{Secs.}
\Crefname{section}{Section}{Sections}
\Crefname{table}{Table}{Tables}
\crefname{table}{Tab.}{Tabs.}

\def\cvprPaperID{9906} 
\def\confName{CVPR}
\def\confYear{2022}

\begin{document}

\title{Fair Contrastive Learning for Facial Attribute Classification}


\author{Sungho Park\textsuperscript{1} \quad Jewook Lee\textsuperscript{1} \quad Pilhyeon Lee\textsuperscript{1} \quad Sunhee Hwang\textsuperscript{2} \quad Dohyung Kim\textsuperscript{3} \quad Hyeran Byun\textsuperscript{1}\thanks{Corresponding author.}  \\ \\
\textsuperscript{1}Yonsei University \quad \textsuperscript{2}LG Uplus \quad \textsuperscript{3}SK Inc.
\\{\tt \small \Letter \thinspace \thinspace qkrtjdgh18@yonsei.ac.kr}
%
}


\maketitle

\begin{abstract}
Learning visual representation of high quality is essential for image classification. Recently, a series of contrastive representation learning methods have achieved preeminent success. Particularly, \textit{SupCon}~\cite{supcon} outperformed the dominant methods based on cross-entropy loss in representation learning. However, we notice that there could be potential ethical risks in supervised contrastive learning. In this paper, we for the first time analyze unfairness caused by supervised contrastive learning and propose a new Fair Supervised Contrastive Loss (\textit{FSCL}) for fair visual representation learning. Inheriting the philosophy of supervised contrastive learning, it encourages representation of the same class to be closer to each other than that of different classes, while ensuring fairness by penalizing the inclusion of sensitive attribute information in representation. In addition, we introduce a group-wise normalization to diminish the disparities of intra-group compactness and inter-class separability between demographic groups that arouse unfair classification. Through extensive experiments on CelebA and UTK Face, we validate that the proposed method significantly outperforms \textit{SupCon} and existing state-of-the-art methods in terms of the trade-off between top-1 accuracy and fairness. Moreover, our method is robust to the intensity of data bias and effectively works in incomplete supervised settings.
Our code is available at \url{https://github.com/sungho-CoolG/FSCL}.

\end{abstract}


\section{Introduction}\label{intro}

Learning powerful visual representation is important for reliable performance in image classification. For a long time, most work has relied on cross-entropy loss to learn the representation due to its strong performance~\cite{resnet,VGG,randaugment,selftraining}.
Meanwhile, recent studies based on contrastive learning have been bringing a new paradigm for representation learning~\cite{memorybank,moco,simclr,mutualcontrastive,multiview}. They effectively learn visual representation by drawing positive pairs and pushing away negative ones in the high-dimensional space.
Despite being originally introduced for unsupervised learning, the contrastive learning strategy proves to be effective in various vision fields~\cite{lee2021completeness,jeon2021stylizationDG,lee2022intersubject}.
Particularly, \textit{SupCon}~\cite{supcon} achieves better top-1 accuracy than the state-of-the-art methods based on the cross-entropy loss on ImageNet~\cite{imagenet} by simply grafting the contrastive loss to the supervised representation learning. 

In this paper, we point out that the contrastive loss may pose potential ethical risks. Despite exhibiting strong performance, it has been underexplored in consideration of fairness which means that the outputs from a model should not be discriminatory in terms of sensitive attributes, such as ethnicity, gender, and age. It is a crucial ethical topic and should be diagnosed in order for the model to be leveraged in the real world~\cite{gorilla,faceapp}. To this end, we analyze the representative contrastive learning model (\textit{SupCon}) from two major perspectives causing unfairness and propose a new contrastive loss to address both of them.

Learning sensitive attribute information is one of the principal causes of unfairness~\cite{FFVAE,controll,ondisentangle,accv_ours}. It incurs unfair classification by inducing a classifier to determine a decision boundary based on undesirable grounds (\textit{i.e.}, sensitive attributes)~\cite{lff,lnl}. From this point of view, we demonstrate that learning sensitive attribute information leads to the decrease of \textit{SupCon} loss on the biased dataset, although the desired behavior is to exclusively learn target class information. Consequently, a model learns both kinds of information to minimize the loss, which eventually aggravates unfairness.

To solve the problem, we propose a Fair Supervised Contrastive Loss (\textit{FSCL}) which prevents encoder networks from learning sensitive attribute information. Basically, it inherits the spirit of supervised contrastive learning that encourages an anchor to be more similar to samples of the same class (\textit{i.e.}, positive samples) than those of other classes (\textit{i.e.}, negative samples). Simultaneously, we limit negative samples to only those having the same sensitive attribute with the anchor among them. In this way, we ensure that learning sensitive attribute information no longer helps the contrastive learning. Rather, it hinders optimizing the loss by increasing the similarity between the anchor and negative samples.

On top of that, we analyze \textit{SupCon} in terms of data imbalance between demographic groups, which is another causal factor of unfairness~\cite{FairGen}. Concretely, we identify that the imbalanced number of anchors and positive samples between the demographic groups encourages the \textit{SupCon} loss to put more weight on majority groups.
As a result, samples from the majority groups generally have higher similarity to the other samples within the same group and lower similarity to samples having different target classes compared to those from the minority groups. We call the former intra-group compactness and the latter inter-class separability. Since their disparities between the groups result in imbalanced classification performances~\cite{intraclass1,intraclass2,intraclass3}, we introduce a group-wise normalization that reduces the gaps by balancing the loss based on the cardinality of anchors and positive samples between the groups.
In the experiments, we demonstrate that it further improves fairness with little damage to the classification performance.

To validate the effectiveness of our method, we perform facial attribute classification on CelebA~\cite{celeba} and UTK Face~\cite{utkface} datasets. In various scenarios, the proposed method significantly ameliorates fairness over \textit{SupCon} and outperforms the state-of-the-art methods in terms of the trade-off between classification accuracy and fairness. Besides, our method is robust to the intensity of data bias and effectively improves fairness even in incompletely supervised settings (\textit{e.g.}, without target class labels or with only a few sensitive attribute labels). Furthermore, we show the extensibility of our method to general bias mitigation through experiments on Dogs and Cats dataset~\cite{Dogs}.

\textbf{Main contributions.} Our main contributions are summarized as follows. 1)~We analyze the causes of unfairness in contrastive learning and propose a Fair Supervised Contrastive Loss that improves fairness by penalizing the inclusion of sensitive attribute information in representation. 2)~We introduce a group-wise normalization, which mitigates the group-wise disparities of intra-group compactness and inter-class separability that exacerbate unfairness of representation. 3)~Through extensive experiments, we validate that our method learns fair representation under various environments. It achieves the best trade-off performances between top-1 accuracy and fairness on CelebA and UTK Face.

\section{Related Work}
\subsection{Fair Representation Learning}
\label{fairness}

Several studies~\cite{grl,mitigating_unwanted,unwanted2,lnl} tried to learn fair representation through adversarial training.
They adversarially train the encoder network and the classification head for sensitive attributes so that the encoder network is agnostic to sensitive attribute information.~\cite{grl} learned fair representation by reversing gradients of classification loss for sensitive attributes through gradient reversal layer (GRL).~\cite{lnl} further minimized the mutual information between representation and sensitive attribute labels to eliminate their correlations.~\cite{mitigating_unwanted,unwanted2} designed structures in which the outputs from the classification head for target classes are fed into that for sensitive attributes. The latter head removes bias for sensitive attributes in the intermediate outputs through GRL.


Disentangled representation learning~\cite{orthogonal,FFVAE,FDVAE} is another mainstream for fair representation learning.~\cite{orthogonal} enforced two types of representation respectively for target classes and sensitive attributes to be orthogonal to each other by maximizing the entropy of the opposite information in each representation.~\cite{FFVAE} leveraged the disentanglement loss~\cite{factorvae} to separate the representation space into sensitive latents and non-sensitive latents without target class labels. Both methods improved fairness by discarding representation containing sensitive attribute information in downstream classification. Moreover,~\cite{FDVAE} pointed out the shortcoming of~\cite{FFVAE} that information related to both target and sensitive attributes is contained in sensitive latents and discarded. They introduced an additional subspace for the intersected information.

Recently,~\cite{MFD} made a fresh attempt to improve fairness without compromising performance through fair knowledge distillation. Based on MMD~\cite{MMD}, they encourage the feature distribution of the student model conditioned by sensitive attributes to get close to that of the teacher model averaged over the sensitive attributes. With an oversampling strategy, they ameliorate both classification accuracy and fairness on the balanced test set. Meanwhile,~\cite{FairGen} proposed a perturbation method which decorrelates the target and sensitive attributes in the latent space of a pre-trained GAN. Then, they generate a balanced dataset with it and utilized the dataset for a fair training of a classification network.

\subsection{Contrastive Representation Learning}

Contrastive learning~\cite{memorybank,multiview,pretextinvariant,moco,simclr} has become a dominant approach to learning visual representation in a self-supervised manner. Without class labels, they learned outstanding representation by pulling samples from the same image together and pushing away those from different images.~\cite{memorybank,multiview,pretextinvariant} indicated that the number of negative samples is important for contrastive learning and introduced memory banks to 
increase it without exploding GPU memory consumption. To solve the inconsistency problem between the updated encoder networks and outdated memory bank,~\cite{moco} utilized a dynamic memory queue as a memory bank and updated it with a slowly moving momentum encoder. Furthermore,~\cite{simclr} proposed a simple architecture for contrastive learning (\textit{i.e.}, \textit{SimCLR}) that outperforms previous methods without the memory bank and specialized architectures. Based on it,~\cite{supcon} proposed a supervised version of contrastive loss (\textit{i.e.}, \textit{SupCon}). Unlike the previous methods, they set all samples having the same class with an anchor to positive samples and pull them to the anchor.

\section{Method}\label{method}

In this section, we first analyze the causes of unfairness in supervised contrastive learning, and then describe the proposed method to solve them. Our method is based on a simple framework for contrastive learning similar to previous works~\cite{simclr,supcon}. We note that our key contributions lie in not introducing a specific framework but designing a new general loss for learning fair and informative representation.


\subsection{Preliminaries}
\subsubsection{Overall flow}
Assume that we have randomly sampled $N$ data pairs in a batch, $\{x_k,y_k,s_k\}_{k=1...N}$. Here, $x_k \in X$, $y_k \in Y$, and $s_k\in S$ respectively denote an input image, its target class label out of $N_y$ classes, and its sensitive attribute label out of $N_s$ classes. Following the prior works~\cite{simclr,moco}, we randomly crop each image $x_k$ to generate two independent patches (\ie, views), $\hat{x}_{2k-1}, \hat{x}_{2k} \in \hat{X}$, resulting in a multi-view batch, $\{\hat{x_l},\hat{y}_l,\hat{s}_l\}_{l=1...2N}$, where $\hat{y}_{2k-1}=\hat{y}_{2k}$ and $\hat{s}_{2k-1}=\hat{s}_{2k}$ for $k \in [1, N]$. An encoder network $\mathcal{F}(\cdot)$ maps the image patches into representation $H=\{h_{l}\}_{l=1...2N}$, then a projection network $\mathcal{G}(\cdot)$ in turn maps $h_l$ into another representation $Z=\{z_l\}_{l=1...2N}$ for contrastive learning. The encoding networks (\textit{i.e.}, $\mathcal{F}$ and $\mathcal{G}$) are jointly optimized with contrastive objectives and this process is called representation learning. 

After the representation learning process, we freeze the encoder network and throw away the projection network. The frozen encoder network produces representation $h_k$ from the input image $x_k$ instead of the cropped views. Taking the representation as input, a classifier is trained to predict the target class label $y_k$ using the cross-entropy loss.

\subsubsection{Self-supervised and supervised contrastive losses}

Both self-supervised and supervised contrastive loss enforce an anchor to be more similar to positive samples than negative samples. The major difference is the way positive and negative samples are defined. In the self-supervised version~\cite{simclr}, for an anchor $\hat{x}_i$, the other view from the same image is defined as the positive sample. Meanwhile, in the supervised version~\cite{supcon}, all patches sharing the same target class labels with the anchor $\hat{x}_i$ are assigned to positive samples, \ie, $\hat{X}_p(i)=\{\hat{x}_p \in \hat{X}|\hat{y}_p=\hat{y}_i,\hat{x}_p \neq \hat{x}_i\}$. In both settings, patches that are neither positive samples nor the anchor are set to negative samples, \ie, $\hat{X}_n(i)=\{\hat{x}_n \in \hat{X}|\hat{x}_n \notin \hat{X}_p(i),\hat{x}_n \neq \hat{x}_i\}$.

In the latent space of $z_l$, the self-supervised loss maximizes the log-softmax of the similarity between $z_i$ and $z_p$ for the similarity between $z_i$ and representation of all the other samples, $\hat{X}_a(i)=\hat{X}_p(i) \cup \hat{X}_n(i)$. The supervised loss calculates the normalized summation of the multiple log-softmax for all $z_p$ and maximizes it. The self-supervised contrastive loss ($L^{SS}$) and supervised contrastive loss ($L^{Sup}$) are formulated as follows.  
\begin{equation}
   L^{SS}=-\sum_{z_i\in Z}\log \frac{\phi_{p}}{\sum_{z_a\in Z_a(i)}\phi_{a}},
\end{equation}          
\begin{equation}
\label{eq:sup_con}
    L^{Sup}=-\sum_{z_i\in Z}\frac{1}{\left | Z_p(i) \right |}\sum_{z_p\in Z_p(i)}\log\frac{\phi_{p}}{\sum_{z_a\in Z_a(i)}\phi_{a}},
\end{equation} 
where $\phi_{x}$ denotes $exp(z_i\cdot z_{x}/\tau)$, $x \in \{a,p\}$. $\tau$ is a temperature parameter, which is set to lower than 1 for sharper distribution of the softmax scores. $|Z_p(i)|$ is the number of positive samples for an anchor $z_i$. In $L^{Sup}$, the cardinality of positive samples varies from anchor to anchor and the factor $\frac{1}{|Z_p(i)|}$ normalizes it. The multiple positive samples and normalization factor ensure that $L^{Sup}$ achieves better classification performances than $L^{SS}$.

\begin{figure*}[t]

  \centering
  \includegraphics[clip=true, width=0.90\textwidth]{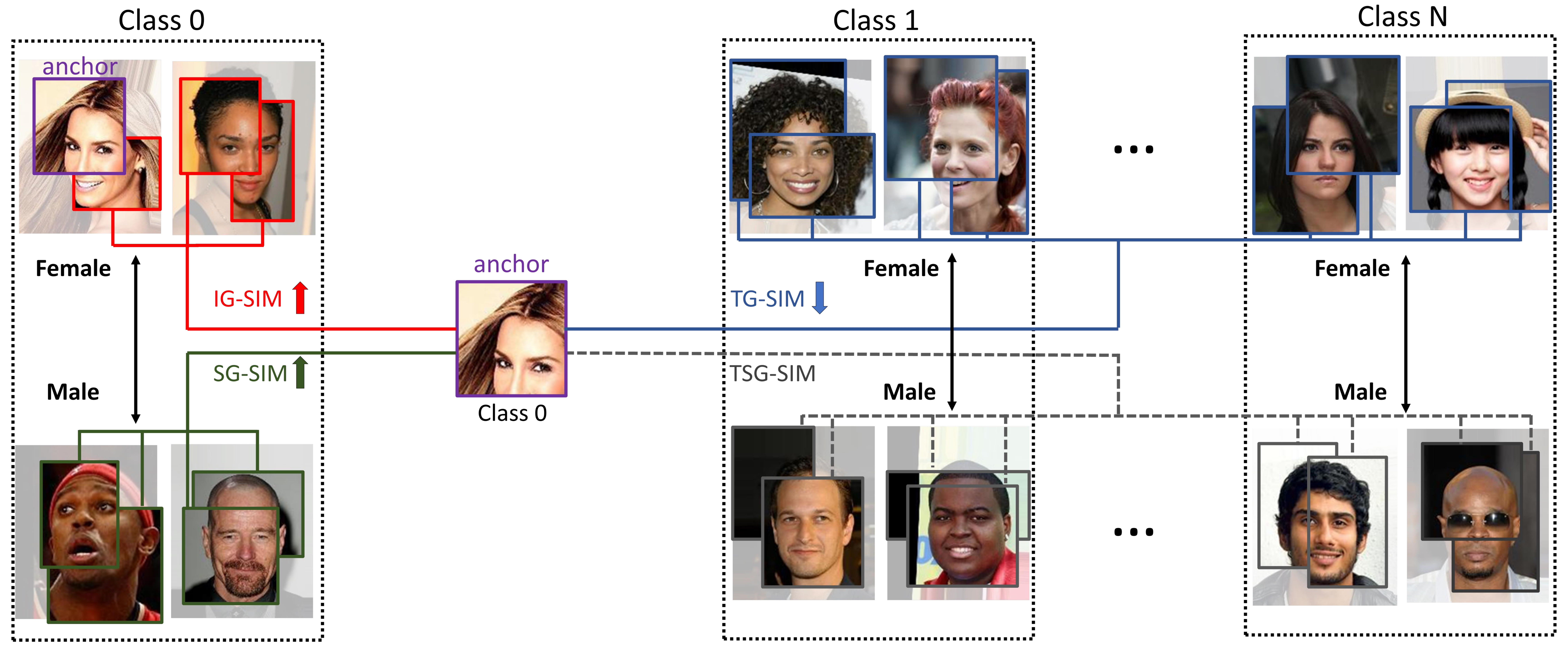}
  \caption{\textbf{The concept of fair supervised contrastive loss (\textit{FSCL}).} It increases the similarity between an anchor and samples of the same target class (IG-SIM and SG-SIM). On the contrary, it decreases the similarity between the anchor and samples having different target classes and the same sensitive attribute, \textit{e.g.,} gender, (TG-SIM). Here TSG-SIM is not directly affected by the loss.}
  \label{model}
\end{figure*}

\subsection{Unfairness in Supervised Contrastive Loss}
\subsubsection{Learning of sensitive attribute information} 
\label{assumption}
As revealed in the literature~\cite{FFVAE,controll,ondisentangle}, learning sensitive attribute information is one of the key factors causing unfair classification. Therefore, to analyze the unfairness of $L^{Sup}$, we start by exploring whether the loss encourages encoder networks to learn the malignant information.


Specifically, we define learning of sensitive attribute information as increasing $I(Z;S)=\mathbb{E}_{P(z,s)}\log \frac{P(z,s)}{P(z)P(s)}$~\cite{lnl,FFVAE}, which is mutual information between $Z$ and $S$. Subsequently, we suppose two random points, $t_l$ and $t_m$, in training time, where $I(Z;S)$ is higher at $t_m$ than at $t_l$ (Assumption 1). In addition, to simplify a wide variety of and complicated data bias, we defined an ideally biased dataset, $\{\tilde{X}$, $\tilde{Y}$, $\tilde{S}$\}, where each target attribute is correlated with one different sensitive attribute in equal intensity. We provide further details on it in Appendix. Then we demonstrate that $L^{Sup}$ will lead the encoding networks to learn sensitive attribute information by proving the theorem below.

\noindent $~~$\textbf{Theorem 1} $~$Given $\tilde{X}$, $\tilde{Y}$, and $\tilde{S}$, for all $ t_l,~t_m $, $V^{t_l}>V^{t_m}$. 

Here, $V^{t_l}$ and $V^{t_m}$ denote the values of $L^{Sup}$ at $t_l$ and $t_m$, respectively. Theorem 1 represents that the value of $L^{Sup}$ is always larger at $t_l$ than at $t_m$. In other words, $L^{Sup}$ is inversely proportional to the $I(Z;S)$. Therefore, it results in the following Corollary. Due to the space limit, we provide the mathematical proof in Appendix.



\noindent $~~$\textbf{Corollary 1} ~Learning sensitive attribute information decreases $L_{sup}$, given $\tilde{X}$, $\tilde{Y}$, and $\tilde{S}$.

In conclusion, both learning the target attribute and sensitive attribute information reduce $L^{Sup}$. Since the encoding networks do not have the intrinsic ability to distinguish them, they will learn both kinds of information to optimize $L^{Sup}$, which eventually aggravates unfairness. 

\subsubsection{Data imbalance between demographic groups}\label{group-wise imbalance}

In addition to discovering the malignant information learning, we explore another cause of unfairness with respect to the imbalanced number of data between data groups. Concretely, we define a data group as a bundle of data having common target classes and sensitive attributes. Based on the group definition, we reformulate \textit{$L^{Sup}$} as follows.
\begin{equation}
\begin{split}
     L^{Sup}=-&\sum_{\forall j,k}\sum_{z_i\in Z^{j, k}}\frac{1}{\left | Z_p(i) \right |}  \\&\sum_{\forall k} \sum_{z_p\in Z_{p}^{k}(i)}  \log \frac{\phi_{p}}{\sum_{z_{a}\in Z_{a}(i)}\phi_{a}},
\end{split}
\label{eq:sup_con_reform}
\end{equation}
where $Z^{j, k}=\{z_i \in Z | \hat{y}_i=j, \hat{s}_i=k  \}$ and $Z_{p}^{k}(i)=\{z_{p} \in Z_{p}(i) | \hat{s}_p=k\}$ for $j \in [1, N_y],$ and $k \in [1, N_s]$. We note that Eq.~\ref{eq:sup_con} and \ref{eq:sup_con_reform} are mathematically identical, but the latter reveals that the imbalanced number of anchors between $Z^{j, k}$ and that of positive samples between $Z_{p}^{k}(i)$ are not normalized by the existing factor $1/|Z_p(i)|$. It results in the loss putting more weight on the majority groups, and thus the loss encourages the majority groups to have better intra-group compactness and inter-class separability compared to the minority groups. Consequently, as indicated in ~\cite{intraclass1,intraclass2,intraclass3}, their disparities between the groups incur unfair classification performances.





\subsection{Fair Supervised Contrastive Loss}

To resolve the problem of learning sensitive attribute information (Sec.~\ref{assumption}), we propose a Fair Supervised Contrastive Loss (\textit{FSCL}) which explicitly penalizes that the encoding networks learn the unwanted information. For brief and clear explanations, we define the following similarities in consideration of the relationship between an anchor and the other samples. 
\begin{itemize}
     \item \textbf{IG-SIM} (Intra-Group Similarity) is the similarity between an anchor and samples within the same group, which have the \textit{same} target class and the \textit{same} sensitive attribute with the anchor.
     The sample set is defined as $Z_{ig}(i)=\{z_{ig} \in Z_{p}(i) | \hat{s}_{ig} = \hat{s}_{i}, \hat{y}_{ig} = \hat{y}_{i}\}$.
     
    \item \textbf{SG-SIM} (Sensitive Inter-Group Similarity) is the similarity between an anchor and samples that belong to groups with the \textit{same} target class and \textit{different} sensitive attributes with the anchor.
    The sample set is defined as $Z_{sg}(i)=\{z_{sg} \in Z_p(i) | \hat{s}_{sg} \neq \hat{s}_{i} , \hat{y}_{sg} = \hat{y}_{i}\}$.
    \item \textbf{TG-SIM} (Target Inter-Group Similarity) is the similarity between an anchor and samples that belong to groups with \textit{different} target classes and the \textit{same} sensitive attribute with the anchor.
    The sample set is defined as $Z_{tg}(i)=\{z_{tg} \in Z_{n}(i) | \hat{s}_{tg} = \hat{s}_{i}, \hat{y}_{tg} \neq \hat{y}_{i} \}$)
    \item \textbf{TSG-SIM} (Target \& Sensitive Inter-Group Similarity) is the similarity between an anchor and samples that belong to groups with \textit{different} target classes and \textit{different} sensitive attributes with the anchor.
    The sample set is defined as \textit{i.e.}, $Z_{tsg}(i)=\{z_{tsg} \in Z_{n}(i) | \hat{s}_{tsg} \neq \hat{s}_{i}, \hat{y}_{tsg} \neq \hat{y}_{i} \}$.
    
   
\end{itemize}
Our key idea is to define the negative sample set as the samples with the same sensitive attributes and different target classes with the anchor (\textit{i.e.}, $Z_{tg}$). Based on this, we design \textit{FSCL} that encourages IG-SIM and SG-SIM to be higher than TG-SIM, as illustrated in Figure~\ref{model}. It is formulated as follows.

\begin{equation}\label{fscl}
    FSCL=-\sum_{z_i\in Z}\frac{1}{\left | Z_p(i) \right |}\sum_{z_p\in Z_p(i)}\log \frac{\phi_{p}}{\sum_{z_{tg}\in Z_{tg}(i)}\phi_{tg}}, 
\end{equation}
where $Z_p(i)= Z_{ig}(i) \cup Z_{sg}(i) $ and $\phi_{tg}=exp(z_i\cdot z_{tg}/\tau)$.

On a case-by-case basis, we explain how our \textit{FSCL} addresses the problem of learning unwanted information. In a case of $z_p \in Z_{ig}(i)$, the positive samples and negative samples (\textit{i.e.}, $ Z_{tg}(i)$) all have the same sensitive attributes with the anchor. Therefore, the encoding networks no longer consider the sensitive attribute information to be a valuable feature for contrasting an anchor with the negative samples more than with the positive samples.

In the other case of $z_p \in Z_{sg}(i)$, the positive samples have different sensitive attributes from the anchor and negative samples (\textit{i.e.}, $ Z_{tg}(i)$). If the encoding networks learn the sensitive attribute information, the similarity between the positive samples and the anchor (\textit{i.e.}, $\phi_{p}$) will decrease and the similarity between the negative samples and it (\textit{i.e.}, $\phi_{tg}$) will increase, which is contrary to the objective of the loss. As a result, minimizing the loss inhibits learning the unwanted information in this case.


\subsection{Group-wise Normalization}

As aforementioned in Sec.~\ref{group-wise imbalance}, the imbalanced number of anchors and positive samples between data groups causes the group-wise disparities in terms of intra-group compactness and inter-class separability. To alleviate the unfairness brought by the disparities, we introduce group-wise normalization as follows.
\begin{equation}
\begin{split}
    FSCL\text{+}=-\sum_{\forall j,k}\frac{1}{\left | Z^{j, k} \right |}&\sum_{z_i\in Z^{j, k}}\sum_{\forall k} \frac{1}{\left | Z_{p}^{k}(i) \right |} \\\sum_{z_p\in Z_{p}^{k}(i)}&\log \frac{\phi_{p}}{\sum_{z_{tg}\in Z_{tg}(i)}\phi_{tg}},
\end{split}
\end{equation}
where $1/|Z^{j, k}|$ and $1/|Z_{p}^{k}(i)|$ are the group-wise normalization factors. Different from the existing factor in $L^{Sup}$ (\textit{i.e.,} $1/|Z_p(i)|$), they normalize the cardinality of anchors and positive samples within each group. On an experimental basis, we demonstrate that the proposed normalization mitigates the group-wise imbalances in terms of intra-group compactness and inter-class separability.

\begin{table*}[t]
\centering
\resizebox{0.99\textwidth}{!}{
\begin{tabular}{ccccccccccccccccccccccccccc}
\toprule
\multirow{2}{*}{Method } & \multicolumn{2}{c}{T=\textit{a} / S=\textit{m}} && \multicolumn{2}{c}{T=\textit{a} / S=\textit{y}} && \multicolumn{2}{c}{T=\textit{b} / S=\textit{m}} && \multicolumn{2}{c}{T=\textit{b} / S=\textit{y}} && \multicolumn{2}{c}{T=\textit{e} / S=\textit{m}} && \multicolumn{2}{c}{T=\textit{e} / S=\textit{y}} && \multicolumn{2}{c}{T=\textit{a} \& \textit{o} / S=\textit{m} } && \multicolumn{2}{c}{T=\textit{e} \& \textit{b} / S=\textit{m}} && \multicolumn{2}{c}{T=\textit{a} / S=\textit{m} \& \textit{y}} \\ \cmidrule[0.5pt]{2-3} \cmidrule[0.5pt]{5-6} \cmidrule[0.5pt]{8-9} \cmidrule[0.5pt]{11-12} \cmidrule[0.5pt]{14-15} \cmidrule[0.5pt]{17-18} \cmidrule[0.5pt]{20-21} \cmidrule[0.5pt]{23-24} \cmidrule[0.5pt]{26-27}
& EO  & Acc. &&  EO   &  Acc.  && EO    &  Acc.  &&  EO    &  Acc.   && EO    &  Acc.   &&  EO  &  Acc.   &&  EO   &  Acc.   &&  EO   &  Acc.   &&  EO   &  Acc.   \\ \cmidrule[0.5pt]{1-27} \morecmidrules\cmidrule[0.5pt]{1-27}
\textit{CE}~\cite{resnet} & 27.8 & 79.6 && 16.8 & 79.8 && 17.6 & 84.0 && 14.7  & 84.5 && 15.0 & 83.9 && 12.7 &  83.8 && 30.0 & 73.9 && 12.9 & 72.6 && 31.3 & 79.5     \\ \cmidrule[0.5pt]{1-27}
\textit{GRL}~\cite{grl} & 24.9 & 77.2 && 14.7 & 74.6 && 14.0 & 82.5 && 10.0 & 83.3 && 6.7  & 81.9 && 5.9 & 82.3 && 17.8  & 73.1 && 9.4 & 71.4 && 22.9 & 78.6   \\ 
\textit{LNL}~\cite{lnl} & 21.8 & 79.9 && 13.7 & 74.3 && 10.7 & 82.3 && 6.8 & 82.3 && 5.0 & 81.6 && 3.3 & 80.3 && 16.7 & 72.9 && 7.4 & 70.8 && 20.7 & 77.7   \\ 
\textit{FD-VAE}~\cite{FDVAE} & 15.1 & 76.9 && 14.8 & 77.5 && 11.2 & 81.6 && 6.7 & 81.7 && 5.7 & 82.6 && 6.2 & 84.0 && 18.2 & 73.4 && 8.2 & 70.2 && 19.9 & 78.0   \\ 
\textit{MFD}~\cite{MFD} & 7.4 & 78.0 && 14.9 & 80.0 && 7.3 & 78.0 && 5.4 & 78.0 && 8.7 & 79.0 && 5.2 & 78.0 && 8.7 & 74.0 && 9.0 & 70.0   && 19.4 & 76.1 \\ \cmidrule[0.5pt]{1-27}
\textit{SupCon}~\cite{supcon} & 30.5 & 80.5 && 21.7 & 80.1 && 20.7 & 84.6 && 16.9 & 84.4 && 20.8 & 84.3 && 10.8 & 84.0 && 22.8  & 74.0 && 12.5 & 72.7 && 24.4 & 81.7   \\ \cmidrule[0.5pt]{1-27}
\textit{FSCL} & 11.5 & 79.1 && 13.0 & 79.1 && 7.0 & 82.1 && 6.4 & 83.8 && 3.8 & 82.7 && 1.8  & 82.0 && 8.1 & 74.1 && 6.8 & 71.1 && 19.9 & 79.4    \\ 
\textit{FSCL+} & \textbf{6.5} & 79.1 && \textbf{12.4} & 79.1 && \textbf{4.7} & 82.9 && \textbf{4.8} & 84.1 && \textbf{3.0} & 83.4 && \textbf{1.6} & 83.5 && \textbf{3.6} & 74.8 && \textbf{2.5} & 70.8 && \textbf{17.0} & 77.2  \\ \bottomrule
\end{tabular}
}
\caption{\textbf{Classification results on CelebA.} We measure classification accuracy (ACC.) and equalized odds (EO) in various scenarios.  Here \textit{a}, \textit{b}, \textit{e}, \textit{o}, \textit{m}, and \textit{y} respectively denote \textit{attractiveness}, \textit{bignose}, \textit{bags-under-eyes}, \textit{mouth-slightly-open}, \textit{male}, and \textit{young}. On the other hand, T and S represent target and sensitive attributes, respectively.
All the results are the averaged scores over three independent runs.
The standard deviations are provided in Appendix.}
\label{table:main}
\end{table*}

\section{Experiment}

\subsection{Datasets}\label{dataset}

\noindent \textbf{CelebA}~\cite{celeba} contains about 200k facial images with 40 binary attribute annotations. We set \textit{male} (\textit{m}) and \textit{young} (\textit{y}) to sensitive attributes and select target attributes having the highest Pearson correlation with the sensitive attributes~\cite{correlation1,correlation2}. Amongst, we manually excluded the extremely correlated attributes for reliable evaluation. For heavy-makeup as example, there are only 22 males with heavy-makeup in test set. As a result, we exploit three single target attributes: \textit{attractiveness} (\textit{a}), \textit{bignose} (\textit{b}), and \textit{bags-under-eyes} (\textit{e}) as well as two pairs of target attributes: $\{$\textit{bignose}, \textit{bags-under-eyes}$\}$ and $\{$\textit{attractiveness}, \textit{mouth-slightly-open} (\textit{o})$\}$.

\noindent \textbf{UTK Face}~\cite{utkface} consists of about 20k facial images with three kinds of annotations: \textit{gender}, \textit{age}, and \textit{ethnicity}. To evaluate fairness in varied levels of data imbalance, we design several imbalanced versions for the training set. Note that the standard protocol on data splits is not provided in this dataset. Concretely, we set \textit{age} and \textit{ethnicity} to the sensitive attributes and \textit{gender} to
the target attribute. \textit{Age} and \textit{ethnicity} are reformed to binary attributes based on whether \textit{age} is under 35 or not and \textit{ethnicity} is Caucasian or not, respectively. A sensitive group (\textit{e.g.}, Caucasian) has male data $\alpha$ times as much as female data and the other sensitive group has the opposite gender ratio. $\alpha$ is set to 2, 3, and 4 to simulate varying bias levels. Unlike the training set, we organize completely balanced validation and test sets for a fair evaluation.

\noindent \textbf{Dogs and Cats}~\cite{Dogs} has 38,500 dog or cat images. In addition to the original species labels (dog or cat), LNL~\cite{lnl} further annotated color labels (bright or dark). We set \textit{color} to the sensitive attribute and \textit{species} to the target attribute.
We compose a \textit{color} biased training set that contains 5 times more black cats than white cats, while 5 times more white dogs than black dogs. For a fair evaluation, we compose the test set to be completely balanced. Note that we utilize this dataset to examine the extensibility of the proposed method to general bias mitigation (\textit{i.e.}, color) beyond its fairness.

\noindent We provide more details on the datasets in Appendix.

\subsection{Fairness Metrics}

A variety of fairness notions are exploited to measure fairness in classification tasks (\textit{e.g.}, demographic parity~\cite{dp2}, equal opportunity, and equalized odds~\cite{odds}). Demographic parity means that the proportion of positive outcomes in each sensitive group should be equal. Although it may be used as reliable metrics in situations where equality of outcome has to be guaranteed, there is a drawback in that a classifier should deliberately misclassify some labels to satisfy it if the proportion of positive outcomes is not equal in the ground truth (GT)~\cite{dp1,odds}. Equal opportunity solves this issue by pursuing the equal true positive rate (TPR) between sensitive groups. However, it does not address unfairness in negative outcomes. In many real-world applications such as facial attribute classification, fairness of positive and negative outcomes is equivalently important. Therefore, equalized odds, which demands both the equal TPR and false positive rate (FPR), are the most suitable to measure fairness in our experiments. Following the definition in~\cite{odds}, we measure the degree of equalized odds (EO) in various settings (\textit{e.g.}, multiple classes or sensitive attributes) as follows.
\begin{equation}
    \underset{\forall y,c,\{s^0,s^1\}\subset S}{\overline{\sum}} \left |  P_{s^0}(C=c~|Y=y)-P_{s^1}(C={}c~|Y=y) \right |,
\end{equation}
where $\overline{\sum}$ is the averaged sum. $y\in Y$ and $c \in C$ are target labels and outputs from a classifier, respectively, and $\{s^0,s^1\}$ is a two-element subset of sensitive attribute groups $S$.

\subsection{Implementation Details}\label{implementation}
For contrastive learning, we utilize ResNet-18~\cite{resnet} for the encoder network $\mathcal{F}$ and a MLP with two hidden layers for the projection network $\mathcal{G}$. The dimensions of latent spaces are set to 256 and 128, respectively. We augment two cropped patches per image following the augmentation strategy in \cite{simclr} and resize them to 128$\times$128. We set the temperature parameter $\tau$ to 0.1 based on the analysis in~\cite{supcon}. We train the encoding networks for 100 epochs in the representation learning stage, and subsequently train the classifier, which is a MLP with one hidden layer, for 10 epochs using the cross-entropy loss. For the experiments with multiple target or sensitive attributes, we combine multiple binary attribute labels into a multi-class label. All comparative models share the same structures of the encoder network and classifier as ours for a fair comparison. The results reported in this paper are averaged over three independent runs. More details for the augmentation strategy, structure of networks, and experiment settings are provided in Appendix.

\subsection{Classification Results on CelebA}

Table~\ref{table:main} shows the classification results on CelebA. For diverse combinations of target and sensitive attributes, we measure classification performances and fairness with top-1 accuracy and equalized odds (EO), respectively. In all the experiments, \textit{Cross-Entropy} (CE)~\cite{resnet} and \textit{SupCon}~\cite{supcon} record excellent top-1 accuracy but suffer from severe unfairness. Notably, the proposed methods (\textit{FSCL} and \textit{FSCL+}) significantly improve EO over them while preserving the competitive performances. Particularly, the comparison between \textit{FSCL}~(blue) and \textit{FSCL+}~(red) shows that the group-wise normalization brings about better fairness while well preserving the performance or even improving it. Furthermore, we compare ours with various state-of-the-art approaches for fairness such as adversarial training (\textit{GRL}~\cite{grl} and \textit{LNL}~\cite{lnl}), disentangled representation learning (\textit{FD-VAE}~\cite{FDVAE}), and fair distillation (\textit{MFD}~\cite{MFD}). \textit{FSCL+} substantially outperforms all the state-of-the-art methods in terms of the trade-off between top-1 accuracy and EO in all the settings. For a clearer comparison of the trade-off performances, we also provide the experimental results in figure form in Appendix.

 \begin{table}[t]
\centering
\resizebox{0.46\textwidth}{!}{
\begin{tabular}{cccc} \toprule
Method & Adversarial Training~\cite{grl} & EO ($\downarrow$) & Acc. ($\uparrow$) \\ \cmidrule[0.5pt]{1-4} \morecmidrules\cmidrule[0.5pt]{1-4}
\multirow{2}{*}{\textit{SupCon}~\cite{supcon}} & \xmark & 30.5$_{\pm 1.3}$ & 80.5$_{\pm 0.7}$ \\ \cmidrule[0.3pt]{2-4}
 & \cmark & 21.0$_{\pm 0.5}$ & 76.6$_{\pm 0.3}$ \\
 
 \cmidrule[0.3pt]{1-4}
\multirow{2}{*}{\textit{FSCL+}} &  \xmark & \textbf{6.5}$_{\pm 0.4}$ & 79.1$_{\pm 0.1}$ \\ \cmidrule[0.3pt]{2-4}
 & \cmark & 9.0$_{\pm 0.5}$ & 79.2$_{\pm 0.1}$ \\ \bottomrule
\end{tabular}
}
\caption{\textbf{Effect of adversarial training in contrastive learning on CelebA dataset.} We utilize GRL~\cite{grl} for the adversarial training. Here $attractiveness$ and $male$ are set to the target class and sensitive attribute, respectively.}
\label{table:grl}
\end{table}

\subsection{Adversarial Training in Contrastive Learning}
Intuitively, to mitigate the unfairness of \textit{SupCon}, one may imagine simply combining adversarial training with it. In Table~\ref{table:grl}, we demonstrate the effect of adversarial training by applying \textit{GRL}~\cite{grl} to \textit{SupCon} and \textit{FSCL+} in the representation learning. For \textit{SupCon}, while improving fairness to an extent, it largely damages the classification performance. In addition, \textit{FSCL+} achieves much better EO and top-1 accuracy than \textit{SupCon} combined with \textit{GRL}.
This indicates that the simple graft of adversarial training to contrastive learning does not sufficiently improve fairness and designing a new method seamlessly integrated into contrastive learning is more effective.
We do not see further improvements in EO when applying \textit{GRL} to FSCL+. 

\subsection{Compatibility with Fair Data Augmentation}
We incorporate our method with Ramaswamy \textit{et al.}~\cite{FairGen}, one of the state-of-the-art pre-processing methods for fair classification. It generates a de-biased dataset through a Progressive GAN~\cite{progressive} and augments the original dataset with the generated one. 
In Table~\ref{table:gan}, we report the performances of the baseline (\textit{i.e., Cross-Entropy}) and \textit{FSCL+} trained on the original/augmented dataset. The results show that \textit{FSCL+} outperforms Ramaswamy \textit{et al.} (2$^{\text{nd}}$ row) in terms of both EO and top-1 accuracy. Besides, the fairness of ours is further enhanced when adopting the fair data augmentation, which indicates its compatibility.


\begin{table}[t]
\centering
\resizebox{0.47\textwidth}{!}{
\begin{tabular}{cccc} \toprule
Method & Ramaswamy \textit{et al.}~\cite{FairGen} & EO ($\downarrow$) & Acc. ($\uparrow$) \\ \cmidrule[0.5pt]{1-4} \morecmidrules\cmidrule[0.5pt]{1-4}
\multirow{2}{*}{\textit{Cross-Entropy}~\cite{resnet}} & \xmark & 27.8$_{\pm 0.2}$ & 79.6$_{\pm 0.5}$ \\ \cmidrule[0.3pt]{2-4}
 & \cmark & 24.1$_{\pm 0.5}$ & 79.6$_{\pm 0.2}$ \\ \cmidrule[0.3pt]{1-4}
\multirow{2}{*}{\textit{FSCL+}} &  \xmark & 6.5$_{\pm 0.4}$ & 79.1$_{\pm 0.1}$ \\ \cmidrule[0.3pt]{2-4}
 & \cmark & \textbf{4.2}$_{\pm 0.4}$ & 79.6$_{\pm 0.1}$ \\ \bottomrule
\end{tabular}
}
\caption{\textbf{Compatibility with fair data augmentation~\cite{FairGen} on CelebA dataset}. We set \textit{attractiveness} and \textit{male} to the target class and sensitive attribute, respectively.}
\label{table:gan}
\end{table}

\begin{figure}[t]
  \centering
  \includegraphics[clip=true, width=0.47\textwidth]{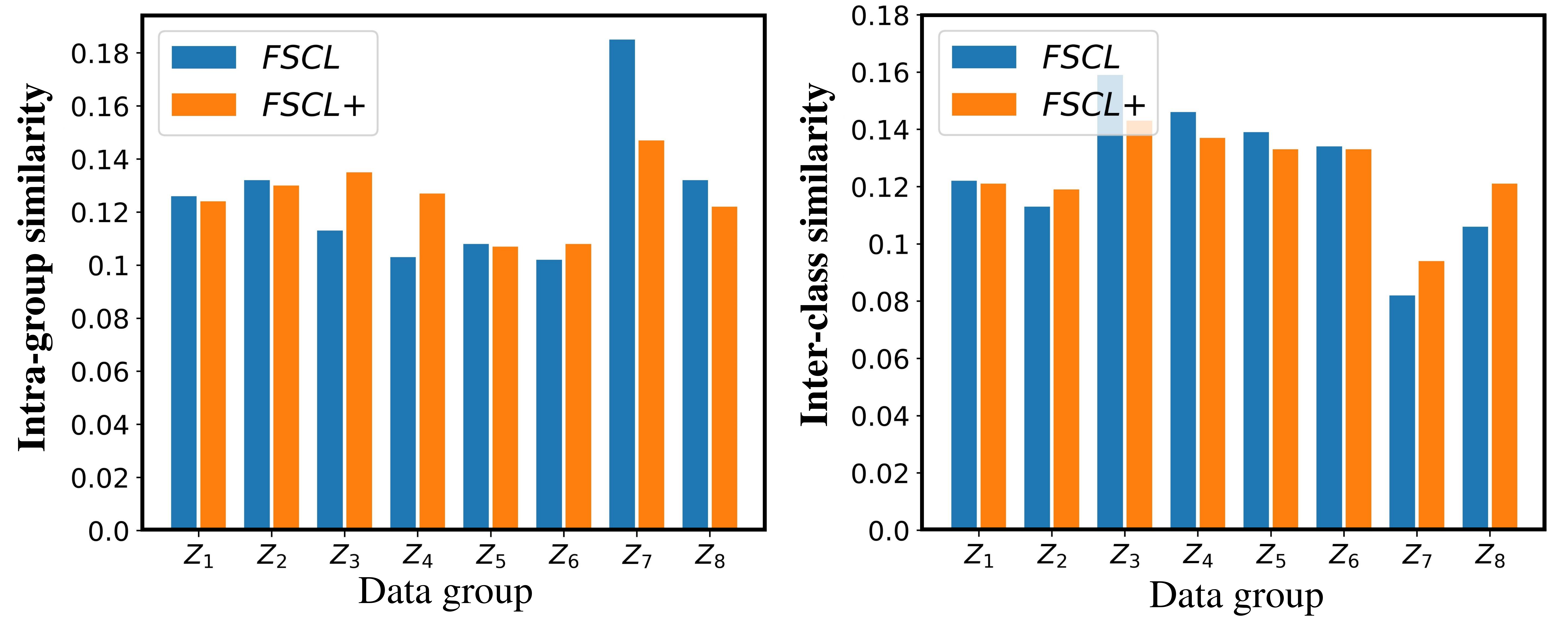}
  \caption{\textbf{Effectiveness of group-wise normalization.} The group-wise normalization (\textit{i.e.}, \textit{FSCL}+) significantly mitigates the group-wise disparities in terms of intra-group and inter-class similarities.}

  \label{fig:ablation}
\end{figure}


\subsection{Effectiveness of Group-wise Normalization}
To analyze the effectiveness of the group-wise normalization, we compare intra-group compactness and inter-class separability between \textit{FSCL} and \textit{FSCL+}. To this end, we first divide the test set into 8 groups with respect to one target class, \textit{attractiveness}, and two sensitive attributes, \textit{male} and \textit{young}, and then calculate them as follows. The former is measured by averaging the similarities between representation within a group (\textit{i.e.}, intra-group similarity) and the latter is measured by averaging the similarity between representation in a group and representation having different class labels with it (\textit{i.e.}, inter-class similarity). For easier comparison, the values are normalized to sum to unity, as shown in Figure~\ref{fig:ablation}. The plots demonstrate that the group-wise normalization significantly diminishes the group-wise disparities. In specific, \textit{FSCL} has the standard deviations of 0.084 and 0.031 in intra-group and inter-class similarities, respectively, while \textit{FSCL+} has lower standard deviations of 0.038 and 0.024.

\begin{figure}[t]
  \centering
  \includegraphics[clip=true, width=0.47\textwidth]{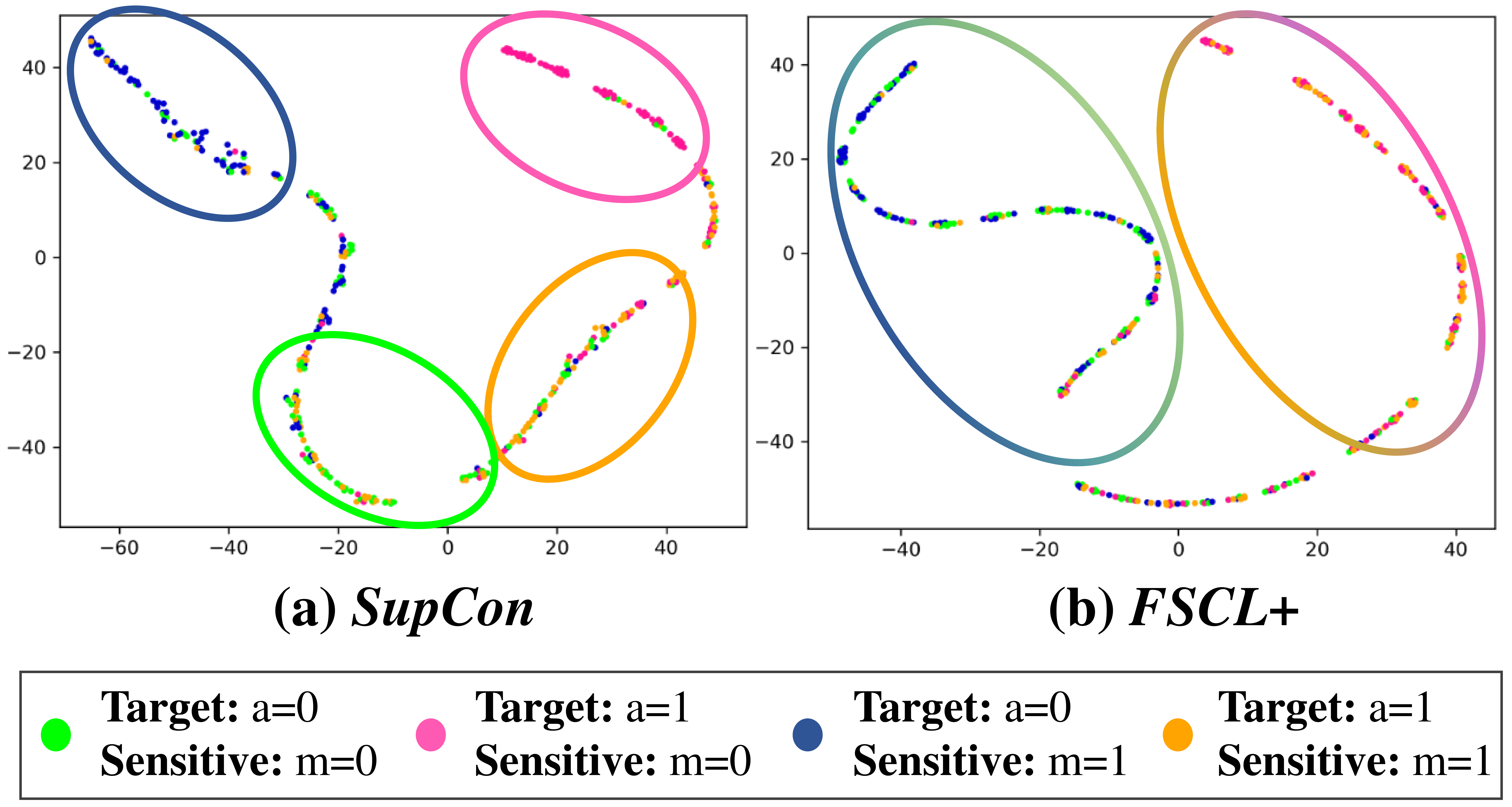}
  \caption{\textbf{Qualitative comparison using t-SNE visualizations.} It is clearly shown that \textit{FSCL+} (b) learns representation more independent of the sensitive attribute than \textit{SupCon} (a).} 
  \label{tsne}
\end{figure}

\subsection{Qualitative Analysis with t-SNE Visualization}

 In Figure~\ref{tsne}, we provide t-SNE plots~\cite{visualtsne} of representation from \textit{SupCon} and \textit{FSCL+} on CelebA dataset. The representation is divided into 4 groups in terms of the target class (\textit{i.e.}, \textit{attractiveness}) and sensitive attribute (\textit{i.e.}, \textit{male}), which are visualized in different colors. In \textit{SupCon}, the representation is divided by both the target class and sensitive attribute, suggesting that the encoding networks learn information for the sensitive attribute as well as the target class. Consequently, the representation of minority groups (\textit{i.e.}, green and orange colors) is more similar to the representation of the counterpart class than that of majority groups (\textit{i.e.}, blue and pink colors). In contrast, in \textit{FSCL+}, the representation is divided by only the target class, that is, it is more agnostic to the sensitive attribute. Accordingly, majority groups can no longer have more privileges than minority groups, which explains why our loss can achieve fairer performance than \textit{SupCon} in image classification. Details of experimental settings are provided in Appendix.

\subsection{Robustness to Severity of Data Bias}
\begin{figure}[t]
  \centering
  \includegraphics[clip=true, width=1\columnwidth]{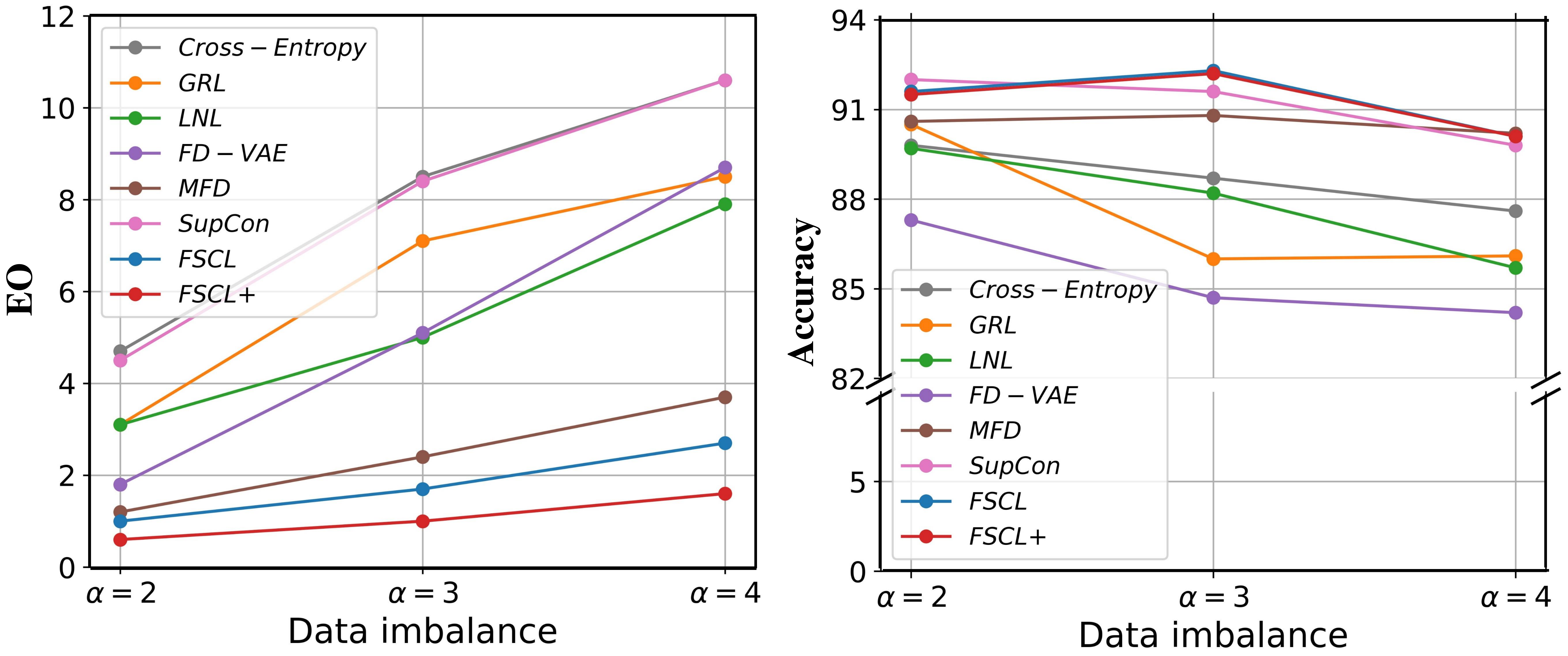}
  \caption{\textbf{Classification results on UTK Face.} We measure classification accuracy and equalized odds (EO) under different severity levels of imbalance ($\alpha$). We set \textit{gender} and \textit{ethnicity} to the target class and sensitive attribute, respectively. Larger $\alpha$ indicates that the training set is more imbalanced.} 
  \label{utkface}
\end{figure}

In Figure~\ref{utkface}, we present the trends of EO and top-1 accuracy according to the intensity shift of data imbalance ($\alpha$) on UTK Face dataset. It can be clearly noticed that our loss best prevents the degradation of fairness caused by an increase in $\alpha$, achieving the fairest performance at all the intensities. In the figure, as $\alpha$ increases, the EO gaps between ours and the others become larger, which manifests the robustness of the proposed methods against the severity of data bias. Moreover, at all the intensities, our loss successfully maintains the top-1 accuracy, which is close to \textit{SupCon}. Experimental results on another sensitive attribute (\ie, \textit{age}) draw similar conclusions and are provided in Appendix.

 \begin{table}[t]
\centering
\resizebox{0.94\columnwidth}{!}{
\begin{tabular}{ccccc} \toprule
\multirow{2}{*}{Method} & \multicolumn{2}{c}{Target:\textit{a} / Sensitive:\textit{m}} & \multicolumn{2}{c}{Target:\textit{b} / Sensitive:\textit{m}} \\ \cline{2-5} 
 & EO ($\downarrow$) & Acc. ($\uparrow$) & EO ($\downarrow$)& Acc. ($\uparrow$)\\ \cmidrule[0.5pt]{1-5} \morecmidrules\cmidrule[0.5pt]{1-5}
\textit{SimCLR}~\cite{simclr} & 29.4$_{\pm 2.5}$ & 75.7$_{\pm 0.2}$ &  16.4$_{\pm 0.4}$ & 82.0$_{\pm 0.1}$ \\ \morecmidrules\cmidrule[0.5pt]{1-5}
\textit{SimCLR}~\cite{simclr}~+~\textit{GRL}~\cite{grl} & 21.9$_{\pm 0.9}$ & 72.3$_{\pm 0.4}$ &13.7$_{\pm 0.3}$ & 82.3$_{\pm 0.0}$\\
\morecmidrules\cmidrule[0.5pt]{1-5}
\textit{FSCL}\textsuperscript{$\dagger$} & \textbf{14.8}$_{\pm 0.9}$ & 74.6$_{\pm 0.4}$ & \textbf{6.1}$_{\pm 0.6}$ & 80.8$_{\pm 0.2}$\\
\bottomrule
\end{tabular}
}

\caption{\textbf{Classification results on CelebA in the absence of target class labels during representation learning.} \textit{FSCL}\textsuperscript{$\dagger$} is a modified version of \textit{FSCL} that does not use the target class labels.}
\label{table:unsupervised}
\end{table}

\subsection{Results in Incomplete Supervised Settings}
We explore a more challenging problem setting, where target class labels are unavailable during the representation learning process.
To this end, we introduce a modified version of \textit{FSCL} that does not exploit target class labels, which is denoted by \textit{FSCL}\textsuperscript{$\dagger$}.
Similar to \textit{SimCLR}~\cite{simclr}, it uses only a single positive sample that comes from the same image with an anchor.
As shown in Table~\ref{table:unsupervised}, ours significantly improves fairness at the acceptable cost of top-1 accuracy, compared to \textit{SimCLR} and \textit{SimCLR+GRL}. Details of the modification are provided in Appendix.



Moreover, we conduct experiments under another challenging environment where only a small portion of data have sensitive attribute labels. One of our simple strategies to handle this task is to generate pseudo-labels for applying \textit{FSCL+} loss. Specifically, we train a classifier to predict sensitive attribute labels only with the samples having sensitive attribute labels, and then generate the pseudo labels of sensitive attributes for the other samples. Another strategy is to apply \textit{FSCL+} loss only to data with sensitive labels and \textit{SupCon} to the other data. 
Table~\ref{table:semi} shows that \textit{FSCL+} effectively ameliorates EO over \textit{SupCon} even under the incomplete supervision of sensitive attributes. 
Surprisingly, \textit{FSCL+} with only 5$\%$ of labels is able to outperform \textit{SupCon+GRL} using all the labels. 



\begin{table}[t]
\centering
\resizebox{0.99\columnwidth}{!}{
\begin{tabular}{ccccc} \toprule
Method & \# of Sensitive& Pseudo-labeling & EO ($\downarrow$) & Acc. ($\uparrow$) \\ \cmidrule[0.5pt]{1-5} \morecmidrules\cmidrule[0.5pt]{1-5}
\textit{SupCon}~\cite{supcon} & 0 & - & 30.5$_{\pm 1.3}$ & 80.5$_{\pm 0.7}$ \\ \morecmidrules\cmidrule[0.5pt]{1-5}
\textit{SupCon}~\cite{supcon}~+~\textit{GRL}~\cite{grl} & 1 & - & 21.0$_{\pm 0.5}$ & 76.6$_{\pm 0.3}$ \\ \morecmidrules\cmidrule[0.5pt]{1-5}
\multirow{9}{*}{\textit{FSCL+}} & 1 & - & \textbf{6.5}$_{\pm 0.4}$ & 79.1$_{\pm 0.1}$ \\ \morecmidrules\cmidrule[0.3pt]{2-5}
 &\multirow{2}{*}{1/2} &\xmark& 13.4$_{\pm 0.1}$ & 79.3$_{\pm 0.3}$ \\
 &  &\cmark& 12.8$_{\pm 1.2}$ & 79.4$_{\pm 0.3}$ \\
 \morecmidrules\cmidrule[0.3pt]{2-5}
 & \multirow{2}{*}{1/4} &\xmark& 18.7$_{\pm 0.3}$ & 80.0$_{\pm 0.3}$ \\
 & &\cmark& 13.4$_{\pm 0.1}$ & 79.5$_{\pm 0.5}$ \\
 \morecmidrules\cmidrule[0.3pt]{2-5}
 & \multirow{2}{*}{1/10} &\xmark& 20.7$_{\pm 0.5}$ &80.2$_{\pm 0.1}$  \\
 & &\cmark& 16.5$_{\pm 0.5}$ &79.6$_{\pm 0.4}$  \\
 \morecmidrules\cmidrule[0.3pt]{2-5}
 & \multirow{2}{*}{1/20} &\xmark& 23.4$_{\pm 0.0}$ & 80.6$_{\pm 0.1}$ \\
 &  &\cmark& 18.8$_{\pm 1.1}$ & 78.5$_{\pm 0.2}$ \\\bottomrule
 
\end{tabular}
}
\caption{\textbf{Classification results on CelebA under incomplete supervision of sensitive attribute labels.} ``\# of Sensitive'' denotes the ratio of data having sensitive attribute labels. We set \textit{attractiveness} and \textit{male} to the target class and sensitive attribute, respectively.}
\label{table:semi}
\end{table}

\begin{figure}[t]
  \centering
  \includegraphics[clip=true, width=0.47\textwidth]{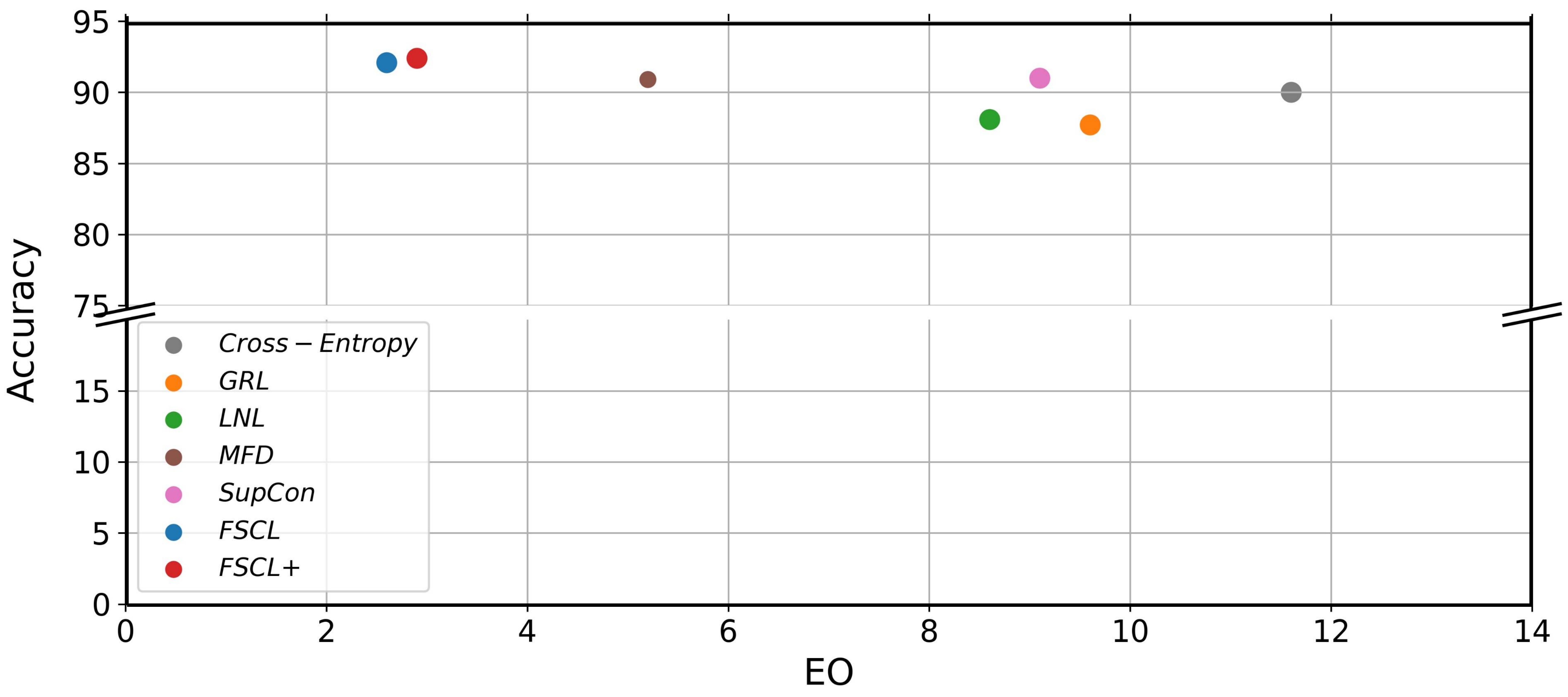}
  \caption{\textbf{Classification results on Dogs and Cats.} We set \textit{species} and \textit{color} to the target and sensitive attributes, respectively.} 
  \label{fig:dogs}
\end{figure}

\subsection{Extensibility to General Bias Mitigation}
To verify the efficacy of the proposed methods in a general bias type, we conduct experiments on Dogs and Cats~\cite{Dogs} with \textit{color} bias. The results are shown in Figure~\ref{fig:dogs}, where our models (\textit{FSCL} and \textit{FSCL+}) best eliminate the color bias, which implies that they are generalizable to various types of bias.
Note that \textit{FSCL}, \textit{FSCL+}, and \textit{MFD} show higher top-1 accuracy than their baselines since fairness improves the performance on the completely balanced test set.

\section{Conclusion}

In this paper, we addressed fairness in contrastive learning. We first analyzed the causative factors of unfairness in the supervised contrastive loss. Then we proposed the fair supervised contrastive loss and introduced the group-wise normalization into the loss. Through extensive experiments, we validated that our loss effectively improves fairness with little degradation of the classification performance.

\subsection*{Acknowledgements}
{
\fontsize{8.4pt}{8.4pt}\selectfont
\noindent This work was supported by the National Research Foundation of Korea (NRF) grant funded by the Korea government (MSIT) (No. 2022R1A2B5B02001467) and Institute for Information \& Communications Technology Planning \& Evaluation (IITP) grant funded by the Korea government (MSIT) (No. 2019-0-01396: Development of Framework for Analyzing, Detecting, Mitigating of Bias in AI model and Training Data, No. 2020-0-01361: Artificial Intelligence Graduate School Program (Yonsei University)).
}
{\small
\bibliographystyle{ieee_fullname}
\bibliography{egbib}
}
\clearpage

\setcounter{section}{0}
\renewcommand\thesection{\Alph{section}}
\section{Definition of Ideally Biased Dataset}

To confine a wide variety of data bias, we first define an ideally biased dataset that satisfies the following conditions.
\begin{enumerate}
    \item The dataset has $m$ target and sensitive classes (\textit{i.e}., $N_t=N_s=m$). Each target and sensitive class contains the same number of data.
    
    \item Target classes are biased to sensitive classes with a one-to-one mapping. That is, each target class has only one \textit{biased sensitive class}, and no more than one target class has the same \textit{biased sensitive class}.
    
    \item In each target class, \textit{biased sensitive class} has $r$ times more data than other sensitive classes.

    \item Target classes are highly biased to sensitive classes (\textit{i.e.}, $r \geq m^2$).

\end{enumerate}

\begin{figure}[h]
  \centering
  \includegraphics[clip=true, width=0.47\textwidth]{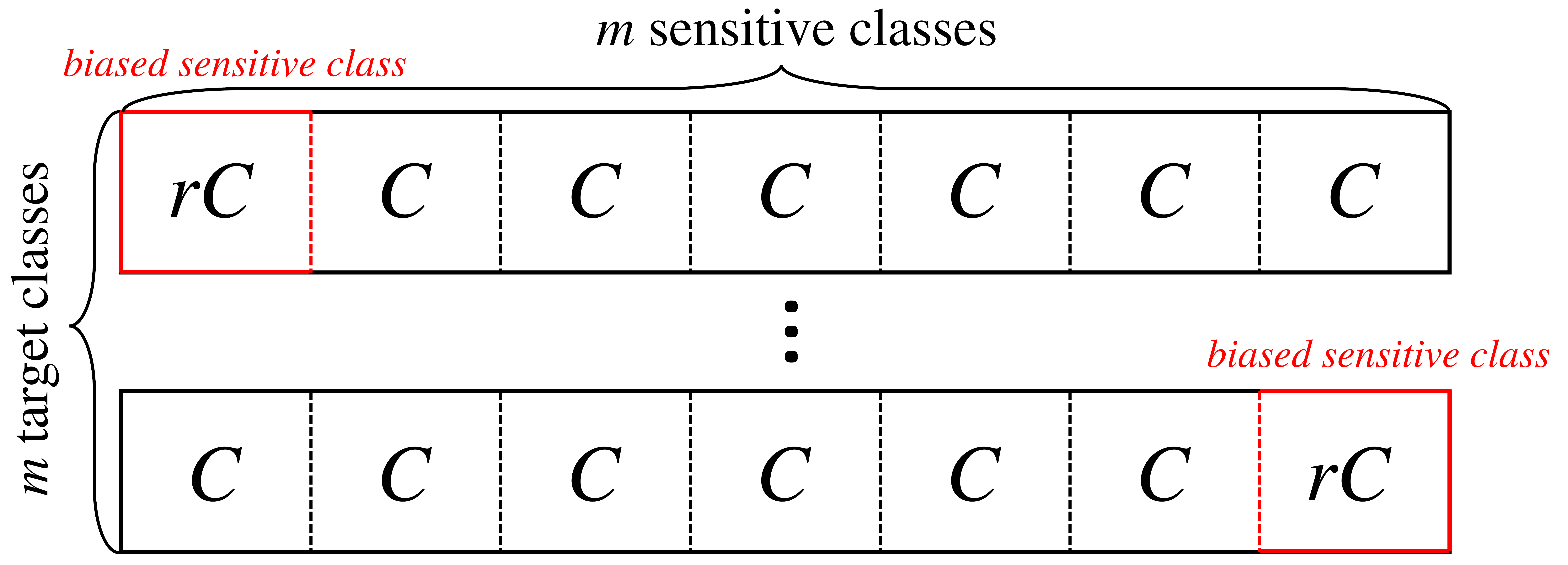} 
  \caption{\textbf{The composition of the ideally biased dataset.} It has $m$ target and sensitive classes, and each target class has only one \textit{biased sensitive class}. $C$ represents the number of data in non-biased classes and $r$ is the bias for \textit{biased sensitive classes}.}
  \label{fig:supple_dataset}
\end{figure}

We illustrated it in Figure~\ref{fig:supple_dataset}, where the number of data in non-biased classes is set to $C$. All the proof below is based on this ideally biased dataset.

\label{ideally}

\section{Mathematical Proof on Theorem 1}

In the main paper, we demonstrated that \textit{SupCon} will lead the encoding networks to learn sensitive attribute information based on Theorem 1. We provide the mathematical proof for the theorem below.
\\

\noindent \textbf{Assumption 1}

Let input data come from the ideally biased dataset (refer to Sec.~\ref{ideally}), where $\tilde{X}$, $\tilde{Y}$, $\tilde{S}$ denote input images, target class labels, and sensitive attribute labels, respectively. We note that target classes are highly correlated with sensitive attributes in the dataset ($r \geq m^2$).
\\



\noindent \textbf{Definition 1}

Learning of sensitive attribute information indicates an increase of $I(Z;\tilde{S})$, where $I(Z;\tilde{S})=\mathbb{E}_{P(z,\tilde{s})}\log \frac{P(z,\tilde{s})}{P(z)P(\tilde{s})}$ and $Z$ denotes the visual representation.
\\


\noindent \textbf{Assumption 2}

Let $t_{l}$, $t_{m}$  be random points in training time when $I(Z^{t_l};\tilde{S})<I(Z^{t_m};\tilde{S})$.
\\

\noindent \textbf{Axiom 1}

Given $\tilde{X}$, $\tilde{Y}$, and $\tilde{S}$, for all $z_i$, $|Z_p(i)|=Cr+(m-1)C-1$, which is a constant.  
\\

\noindent \textbf{Definition 2}\label{def:1}

$L_{a}^{sup} = \sum_{z_i\in Z}\sum_{z_p\in Z_p(i)}\Big[\log\Big(\sum_{z_a\in Z_a(i)} \phi_{a}\Big)\Big]$.  
\\

\noindent \textbf{Definition 3}

$L_{p}^{sup}=\sum_{z_i\in Z}\sum_{z_p\in Z_p(i)} \log  \phi_{p} $.
\\

\noindent \textbf{Proposition 1}

$L^{Sup }= \hat{C}(-L_{p}^{sup}+L_{a}^{sup}$), where $\hat{C}$ is a constant.  
\\

\noindent~~ \textbf{Proof.}
\begin{equation} 
\begin{split}
    L^{Sup}&=-\sum_{z_i\in Z}\frac{1}{\left | Z_p(i) \right |}\sum_{z_p\in Z_p(i)}\log\frac{\phi_{p}}{\sum_{z_a\in Z_a(i)}\phi_{a}}\\
     &=-\sum_{z_i\in Z}\frac{1}{\left | Z_p(i) \right |}\sum_{z_p\in Z_p(i)}\log \phi_{p}\\
    &\quad\quad\quad+\sum_{z_i\in Z}\frac{1}{\left | Z_p(i) \right |}\sum_{z_p\in Z_p(i)}\log\sum_{z_a\in Z_a(i)} \phi_{a}\\
    &=\frac{1}{\left | Z_p(i) \right |}\Big(-L_{p}^{sup}+L_{a}^{sup}\Big)\\
    &=\hat{C}\Big(-L_{p}^{sup}+L_{a}^{sup}\Big) ( \because \mathrm{Axiom 1}).
\end{split}
\end{equation}
\\

\noindent \textbf{Definition 4}

Let $V_x^{t_l}$ and $V_x^{t_m}$ be the values of $L_x^{sup},~x \in \{p,a\}$, at $t_l$ and $t_m$, respectively. 

For example, the value of $L_a^{sup}$ at a certain point in training time, $t_l$, can be represented as:

\begin{equation} 
    V_{a}^{t_l}=\sum_{z_i\in Z}\sum_{z_p\in Z_p(i)} \Biggl[\log \Big(\sum_{z_a\in Z_a(i)}\phi^{t_l}_{a}\Big)  \Biggl],
\end{equation}
where $\phi^{t_k}_{x}=\exp(z_i^{t_k}\cdot z_{x}^{t_k}/\tau),~x \in \{p,a\},~k \in \{l,m\}$.
\\

\noindent \textbf{Definition 5}

Let  $Z_x(i)$=$Z_x^s(i) \cup Z_x^d(i)$, where $Z_{x}^s(i)=\{z_{x} \in Z_{x}(i) | \\ \tilde{s}_x=\tilde{s}_i,\}$ and $Z_{x}^d(i)=$ $\{z_{x} \in Z_{x}(i) |  \tilde{s}_x\neq \tilde{s}_i\}$, $~x \in \{p,a\}$.  
\\

\noindent \textbf{Proposition 2}

From Definition 2, 3, 4 and 5, 
\begin{equation} 
\begin{split}
   V_{a}^{t_k}&=\sum_{z_i\in Z}\sum_{z_p\in Z_p(i)} \Biggl[\log \Big(\sum_{z_a\in Z_a(i)}\phi^{t_k}_{a}\Big)  \Biggl]\\
	&=\sum_{z_i\in Z}\sum_{z_p\in Z_p(i)} \Biggl[\log \Big(\sum_{z_a\in Z_a^s(i)}\phi^{t_k}_{a}
     +\sum_{z_a\in Z_a^d(i)} \phi^{t_k}_{a} \Big) \Biggl].
\end{split}
\end{equation}
\begin{equation} 
\begin{split}
	V_{p}^{t_k}&=\sum_{z_i\in Z}\Biggl[\sum_{z_p\in Z_p(i)} \log \phi^{t_k}_{p}  \Biggl] \\
	&=\sum_{z_i\in Z} \Biggl[\sum_{z_p\in Z_p^s(i)} \log \phi^{t_k}_{p} 
     +\sum_{z_p\in Z_p^d(i)} \log \phi^{t_k}_{p}   \Biggl].
\end{split}
\end{equation}
\\

\noindent \textbf{Conjecture 1}

a) $\sum_{z_a\in Z_a^s(i)}\phi^{t_l}_{a} <\sum_{z_a\in Z_a^s(i)}\phi^{t_m}_{a} $
\\

b) $\sum_{z_a\in Z_a^d(i)}\phi^{t_l}_{a}>\sum_{z_a\in Z_a^d(i)}\phi^{t_m}_{a}$  \\

c) $\sum_{z_p\in Z_p^s(i)}\log\phi^{t_l}_{p}<\sum_{z_p\in Z_p^s(i)}\log\phi^{t_m}_{p}$
\\

d) $\sum_{z_p\in Z_p^d(i)}\log\phi^{t_l}_{p}>\sum_{z_p\in Z_p^d(i)}\log\phi^{t_m}_{p}$  
\\

From Assumption 2, $I(Z^{t_l};\tilde{S})<I(Z^{t_m};\tilde{S})$, hence the similarity between $z_i$ and $Z_k^s(i)$ is larger at $t_m$ than $t_l$. Meanwhile, the similarity between $z_i$ and $Z_k^d(i)$ is smaller at $t_m$ than at $t_l$.
\\

\noindent \textbf{Proposition 3}

Let $\alpha_{z_x}, \beta_{z_x} \in R^+, ~x \in \{p,a\}$, then 
\begin{equation} \label{eq5}
\begin{split}
V_{a}^{t_m}&=\sum_{z_i\in Z}\sum_{z_p\in Z_p(i)} \Biggl[\log \Big(\sum_{z_a\in Z_a^s(i)}(1+\alpha_{z_a})\phi^{t_l}_{a}\\
     &\quad\quad\quad\quad\quad\quad\quad\quad+\sum_{z_a\in Z_a^d(i)} (1-\beta_{z_a})\phi^{t_l}_{a}\Big)  \Biggl],
\end{split}
\end{equation}
\begin{equation} 
\begin{split}
V_{p}^{t_m}&=\sum_{z_i\in Z}\Biggl[\sum_{z_p\in Z_p^s(i)} \log (1+\alpha_{z_p})\phi^{t_l}_{p}\\
     &\quad\quad\quad\quad\quad\quad+\sum_{z_p\in Z_p^d(i)}  \log(1-\beta_{z_p})\phi^{t_l}_{p}  \Biggl],
\end{split}
\end{equation}
where $\alpha_{z_x}$ is an increasing rate of similarity between an anchor and each sample from $t_l$ to $t_m$. Conversely, $\beta_{z_x}$ is the decreasing rate of similarity.  
\\

\noindent~~ \textbf{proof}. 

By Conjecture 1,
\begin{equation} 
\begin{split}
\sum_{z_a\in Z_a^s(i)}\phi^{t_m}_{a}=&\sum_{z_a\in Z_a^s(i)}(1+\alpha_{z_a})\phi^{t_l}_{a},\\
\sum_{z_a\in Z_a^d(i)}\phi^{t_m}_{a}=&\sum_{z_a\in Z_a^d(i)}(1-\beta_{z_a})\phi^{t_l}_{a},\\
\sum_{z_p\in Z_p^s(i)}\log\phi^{t_m}_{p}=&\sum_{z_p\in Z_p^s(i)}\log(1+\alpha_{z_p})\phi^{t_l}_{p},\\
\sum_{z_p\in Z_p^d(i)}\log\phi^{t_m}_{p}=&\sum_{z_p\in Z_p^d(i)}\log(1-\beta_{z_p})\phi^{t_l}_{p}.  
\end{split}
\end{equation}

Therefore,
\begin{equation} 
\begin{split}
V_{a}^{t_m}&=\sum_{z_i\in Z}\sum_{z_p\in Z_p(i)} \Biggl[\log \Big(\sum_{z_a\in Z_a^s(i)}\phi^{t_m}_{a}
     +\sum_{z_a\in Z_a^d(i)} \phi^{t_m}_{a}\Big)  \Biggl]\\ 
&= \sum_{z_i\in Z}\sum_{z_p\in Z_p(i)} \Biggl[\log \Big(\sum_{z_a\in Z_a^s(i)}(1+\alpha_{z_a})\phi^{t_l}_{a}
     \\&\quad\quad\quad\quad\quad\quad\quad\quad\quad\quad\quad
     +\sum_{z_a\in Z_a^d(i)} (1-\beta_{z_a})\phi^{t_l}_{a}\Big)  \Biggl], 
\end{split}
\end{equation}
\begin{equation} 
\begin{split}
	V_{p}^{t_m}&=\sum_{z_i\in Z}\Biggl[\sum_{z_p\in Z_p^s(i)} \log \phi^{t_m}_{p}
     +\sum_{z_p\in Z_p^d(i)}  \log\phi^{t_m}_{p}  \Biggl]\\
	&=\sum_{z_i\in Z}\Biggl[\sum_{z_p\in Z_p^s(i)} \log (1+\alpha_{z_p})\phi^{t_l}_{p}
     \\&\quad\quad\quad\quad\quad\quad\quad+\sum_{z_p\in Z_p^d(i)}  \log(1-\beta_{z_p})\phi^{t_l}_{p}  \Biggl].
\end{split}
\end{equation}
\\

\noindent \textbf{Assumption 3}

Let $\overline{\alpha_{z_x}}$ be the mean increasing rate of similarity (\textit{i.e.}, $\alpha_{z_x}$) over $Z_x^s(i)$ and $\overline{\beta_{z_x}}$ be the mean decreasing rate of similarity (\textit{i.e.}, $\beta_{z_x}$) over $Z_x^d(i)$, then $\overline{\alpha_{z_x}} \approx \overline{\beta_{z_x}}$.
\\

\noindent \textbf{Definition 6}

In Eq.~\ref{eq5}, let the mean $\phi^{t_l}_{a}$ over $Z_a^s(i)$ be $\overline{\phi_{a}^{t_l}}^s$ and that over $Z_a^d(i)$ be $\overline{\phi_{a}^{t_l}}^d$. 
\\

\noindent \textbf{Assumption 4}

Let the difference between $\overline{\phi_{a}^{t_l}}^s$ and $\overline{\phi_{a}^{t_l}}^d$ by sensitive attribute information be $\epsilon \in R^{+}$. Then, $\overline{\phi_{a}^{t_l}}^s \approx \overline{\phi_{a}^{t_l}}^d+\epsilon$, where $\epsilon \ll \overline{\phi_{a}^{t_x}}^d, ~\overline{\phi_{a}^{t_x}}^s$. 
\\



\noindent \textbf{Lemma 1}

Given $\tilde{X}$, $\tilde{Y}$, and $\tilde{S}$, for all $ t_l, t_m $, $V_a^{t_l} \geq V_a^{t_m}$.
\\

\noindent~~ \textbf{proof}. 

From Proposition 3, 
\begin{equation} 
\begin{split}
V_{a}^{t_m}&=\sum_{z_i\in Z}\sum_{z_p\in Z_p(i)} \Biggl[\log \Big(\sum_{z_a\in Z_a^s(i)}(1+\alpha_{z_a})\phi^{t_l}_{a}\\
     &\quad\quad\quad\quad\quad\quad\quad\quad\quad
     +\sum_{z_a\in Z_a^d(i)} (1-\beta_{z_a})\phi^{t_l}_{a}\Big)  \Biggl]\\
        &\approx \sum_{z_i\in Z}\sum_{z_p\in Z_p(i)} \Biggl[\log \Big(\sum_{z_a\in Z_a^s(i)}(1+\overline{\alpha_{z_a}})\phi^{t_l}_{a}
     \\&\quad\quad\quad\quad\quad\quad\quad\quad\quad+\sum_{z_a\in Z_a^d(i)} (1-\overline{\beta_{z_a}})\phi^{t_l}_{a}\Big)  \Biggl].
\end{split}
\end{equation}
Note that $\overline{\alpha_{z_a}}$ and $\overline{\beta_{z_a}}$ are defined in Assumption 3. Then we compare $V_{a}^{t_m}$ and $V_{a}^{t_l}$ as follows.
\begin{equation}\label{eq:compare}
\begin{split}
\Delta V_{a}&=V_{a}^{t_m}- V_{a}^{t_l}\\
&\approx \sum_{z_i\in Z} \sum_{z_p\in Z_p(i)}\Biggl [ \log \Bigg( \frac{\sum_{z_a\in Z_a^s(i)}(1+\overline{\alpha_{z_a}})\phi^{t_l}_{a}}{\sum_{z_a\in Z_a(i)}\phi^{t_l}_{a}}
\\&\quad\quad\quad\quad\quad\quad\quad\quad
+\frac{\sum_{z_a\in Z_a^d(i)} (1-\overline{\beta_{z_a}})\phi^{t_l}_{a}}{\sum_{z_a\in Z_a(i)}\phi^{t_l}_{a}}\Bigg)  \Biggl ]
\\&= \sum_{z_i\in Z} \sum_{z_p\in Z_p(i)} \Biggl  [\log\Bigg(1+\frac{\sum_{z_a\in Z_a^s(i)}\overline{\alpha_{z_a}}\phi^{t_l}_{a}}{\sum_{z_a\in Z_a(i)}\phi^{t_l}_{a}} \\&\quad\quad\quad\quad\quad\quad\quad\quad\quad\quad
-\frac{\sum_{z_a\in Z_a^d(i)} \overline{\beta_{z_a}}\phi^{t_l}_{a}}{\sum_{z_a\in Z_a(i)}\phi^{t_l}_{a}}\Bigg)  \Biggl ].
\end{split}
\end{equation}

By Definition 6, it is rephrased as follows.
\begin{equation}
\begin{split}
    \Delta V_{a}&=\sum_{z_i\in Z} \sum_{z_p\in Z_p(i)}\Biggl [\log\Bigg(1+\frac{\sum_{z_a\in Z_a^s(i)}\overline{\alpha_{z_a}}\,\overline{\phi_{a}^{t_l}}^s}{\sum_{z_a\in Z_a(i)}\phi^{t_l}_{a}}\\ &\quad\quad\quad\quad\quad\quad\quad\quad\quad\quad
    -\frac{\sum_{z_a\in Z_a^d(i)} \overline{\beta_{z_a}}\overline{\phi_{a}^{t_l}}^d}{\sum_{z_a\in Z_a(i)}\phi^{t_l}_{a}}\Bigg)  \Biggl ].
\end{split}
\end{equation}

From Assumption 4, $\overline{\phi_{a}^{t_x}}^s\approx \overline{\phi_{a}^{t_x}}^d+\epsilon$. Based on this, $\Delta V_{a}$ is approximated as follows.


\begin{equation}\label{eq:Va}
\begin{split}
    \Delta V_{a}&\approx\sum_{z_i\in Z} \sum_{z_p\in Z_p(i)}\Biggl[\log\Bigg(1+\frac{\Big(\sum_{z_a\in Z_a^s(i)}\overline{\alpha_{z_a}} }{\sum_{z_a\in Z_a(i)}\phi^{t_l}_{a}}\\&\quad\quad\quad\quad\quad\quad\quad\quad\quad\quad
    \frac{-\sum_{z_a\in Z_a^d(i)} \overline{\beta_{z_a}} \Big) \overline{\phi_{a}^{t_l}}^d }{\quad \quad}\Bigg)  \Biggl],
\end{split}
\end{equation}
where we omit $\epsilon$ for readability since $\epsilon \ll \overline{\phi_{a}^{t_x}}^d, ~\overline{\phi_{a}^{t_x}}^s$.
In the ideally biased dataset, regardless of $z_i$ and $z_p$, $|Z_a^s(i)|$=$rC+(m-1)C$ and $|Z_a^d(i)|=(m-1)rC+(m-1)^2C$. Thus, we can reformulate Eq.~\ref{eq:Va} as follows.
\begin{equation}\label{first}
\begin{split}
     \Delta V_{a}&=\sum_{z_i\in Z} \sum_{z_p\in Z_p(i)}\Biggl [\log\Bigg(1+\frac{\Big((rC+(m-1)C)\overline{\alpha_{z_a}}}{\sum_{z_a\in Z_a(i)}\phi^{t_l}_{a}}\\&\quad\quad\quad\quad
     \frac{-((m-1)rC+(m-1)^2C)\overline{\beta_{z_a}}\Big)\overline{\phi_{a}^{t_l}}^d}{\quad\quad}\Bigg)  \Biggl ]\\
    &=\sum_{z_i\in Z} \sum_{z_p\in Z_p(i)} \Biggl [ \log\Bigg(1+\frac{(m+r-1)C\Big((\overline{\alpha_{z_a}}}{\sum_{z_a\in Z_a(i)}\phi^{t_l}_{a}} \\&\quad\quad\quad\quad\quad\quad\quad\quad\quad\quad\quad
    \frac{-(m-1)\overline{\beta_{z_a}})\overline{\phi_{a}^{t_l}}^d\Big)}{\quad\quad}\Bigg) \Biggl ].
    \end{split}
\end{equation}
 
By Assumption 3, it is approximated as follows.

\begin{equation}
    \begin{split}
        \Delta V_{a}&\approx \sum_{z_i\in Z} \sum_{z_p\in Z_p(i)}\Biggl [ \log\Bigg(1+\frac{(m+r-1)C}{\sum_{z_a\in Z_a(i)}\phi_{a}^{t_l}}\\&\quad\quad\quad\quad\quad\quad\quad\quad\quad \frac{\times \Big(((2-m)\overline{\alpha_{z_a}})\overline{\phi_{a}^{t_l}}^d\Big)}{\quad\quad} \Bigg) \Biggl ]\leq0.
    \end{split}
\end{equation}
Here, $m \geq$ 2, $r > m^{2}$, $C>0$ ($\because$ Assumption 1), $\overline{\alpha_{z_a}}>0$ ($\because$ Proposition 3), and $\phi_{a}^{t_l}>0$ ($\because$ Definition 4).
Therefore, $\Delta V_{a}\leq 0$.\\

\noindent \textbf{Lemma 2}

Given $\tilde{X}$, $\tilde{Y}$, and $\tilde{S}$, for all $ t_l, t_m $, $V_p^{t_l}<V_p^{t_m}$.
\\

\noindent~~ \textbf{proof}.

By Proposition 3, 
\begin{equation}
\begin{split}
    V_{p}^{t_m}&=\sum_{z_i\in Z}\Biggl[\sum_{z_p\in Z_p^s(i)} \log (1+\alpha_{z_p})\phi^{t_l}_{p}\\
    &\quad\quad\quad+\sum_{z_p\in Z_p^d(i)}  \log(1-\beta_{z_p})\phi^{t_l}_{p}  \Biggl]\\
    &\approx\sum_{z_i\in Z} \Biggl[\sum_{z_p\in Z_p^s(i)}\log (1+\overline{\alpha_{z_p}})\phi^{t_l}_{p}\\&
    \quad\quad\quad+\sum_{z_p\in Z_p^d(i)}\log (1-\overline{\beta_{z_p}})\phi^{t_l}_{p}  \Biggl].
\end{split}
\end{equation}

Similar to Eq.~\ref{eq:compare}, we compare $V_{p}^{t_m}$ and $V_{p}^{t_l}$ as follows.
\begin{equation}
    \begin{split}
        \Delta V_{p}=V_{p}^{t_m}- V_{p}^{t_l}&=\sum_{z_i\in Z} \Biggl[\sum_{z_p\in Z_p^s(i)}\log\frac{(1+\overline{\alpha_{z_p}})\phi^{t_l}_{p}}{\phi^{t_l}_{p}}\\&
        \quad\quad\quad+\sum_{z_p\in Z_p^d(i)}\log \frac{(1-\overline{\beta_{z_p}})\phi^{t_l}_{p}}{\phi^{t_l}_{p}}  \Biggl].
    \end{split}
\end{equation}
Here, $\log(1-\overline{\alpha_{z_p}})\approx \log(1-\overline{\beta_{z_p}})$ ($\because$ Assumption 3), and $\log(1-\overline{\alpha_{z_p}}) \approx -\log(1+\overline{\alpha_{z_p}}) $ since $\log(1)=0$ and $\frac{\mathrm{d}\log(1)}{\mathrm{dx}}=1$. Therefore, $\log(1+\overline{\alpha_{z_p}}) \approx -\log(1-\overline{\beta_{z_p}})$. Based on this, we can approximate $\Delta V_{p}$ as follows.
\begin{equation}
    \Delta V_{p}\approx\log(1+\overline{\alpha_{z_p}})\Big (\sum_{z_i\in Z} \sum_{z_p\in Z_p^s(i)}\mathbbm{1}-\sum_{z_i\in Z} \sum_{z_p\in Z_p^d(i)}\mathbbm{1}  \Big),
\end{equation}
where $\mathbbm{1}$ is an indicator function. In the ideally biased dataset, $\sum_{z_i\in Z} \sum_{z_p\in Z_p^s(i)}\mathbbm{1}=(rC)^2+(m-1)C^2$ and $\sum_{z_i\in Z} \sum_{z_p\in Z_p^d(i)}\mathbbm{1}=2(m-1)rC^2+(m-1)(m-2)C^2$. Therefore, we rephrase it as follows.
\begin{equation}
\begin{split}
    &\Delta V_{p}= \Big (\Big((rC)^2+(m-1)C^2\Big)\\&-\Big(2(m-1)rC^2+(m-1)(m-2)C^2\Big) \Big) \log(1+\overline{\alpha_{z_p}})\\
    &= C^2 \Big (  r^2 +(-2m+1)r -m^2+4m-3 \Big )\log(1+\overline{\alpha_{z_p}})\\
    &= C^2 \Big (  (r+\lambda m)(r-((2+\lambda)m-1)) +(4-\lambda)m-3 \Big )\\&\quad\quad\quad \times\log(1+\overline{\alpha_{z_p}})>0 \quad\quad~ \textup{s.t.}\quad~ r>(2+\lambda)m-1
\end{split}
\end{equation}
where $\lambda=-1+\sqrt{2}$. Finally, $\Delta V_{p}>0$ since $m>2$, $r > m^{2}$, and $C>0$ ($\because$ Assumption 1).\\

\noindent~~ \textbf{Theorem 1}

Given $\tilde{X}$, $\tilde{Y}$, and $\tilde{S}$, for all $ t_l, t_m $,  $V^{t_l}>V^{t_m}$.
\\

\noindent~~ \textbf{proof}. 

From Lemma 1 and 2, $V_a^{t_l} \geq V_a^{t_m}$ and $V_p^{t_l}<V_p^{t_m}$ for all $ t_l, t_m $. Since  $V^{t_k}=\hat{C}(-V_p^{t_k}+V_a^{t_k})$ by Proposition 1, $V^{t_l}>V^{t_m}$ for all $ t_l, t_m $.
\\

\noindent \textbf{Corollary 1}

Learning sensitive attribute information decreases $L^{Sup}$, given $\tilde{X}$, $\tilde{Y}$, and $\tilde{S}$.
\\

\noindent~~ \textbf{proof}. 


From Definition 1, learning of sensitive attribute information equals to the increase of $I(Z;\tilde{S})$. In addition, the increase of $I(Z;\tilde{S})$ corresponds to a transition from $t_l$ to $t_m$ since $I(Z;\tilde{S})$ is always higher at $t_m$ than at $t_l$ ($\because$ Assumption 2). Finally, $V^{t_m}$ is always smaller than $V^{t_l}$ ($\because$ Theorem 1), therefore, learning sensitive attribute information decreases $L^{Sup}$.


\begin{table}[h]
\centering
\resizebox{0.38\textwidth}{!}{
\begin{tabular}{cccc} \toprule
Method & Adversarial Training & EO ($\downarrow$) & Acc. ($\uparrow$) \\ \cmidrule[0.5pt]{1-4} \morecmidrules\cmidrule[0.5pt]{1-4}
\textit{SupCon} & \xmark & 30.5$\pm$1.3 & 80.5$\pm$0.7 \\ \cmidrule[0.3pt]{2-4}
 & \cmark & 20.0$\pm$0.3 & 77.2$\pm$0.1 \\ \cmidrule[0.3pt]{1-4}
\textit{FSCL+} &  \xmark & \textbf{6.5$\pm$0.4} & 79.1$\pm$0.1 \\ \cmidrule[0.3pt]{2-4}
 & \cmark & 20.5$\pm$0.4 & 77.8$\pm$0.2 \\ \bottomrule
\end{tabular}
}
\caption{\textbf{Effectiveness of adversarial training in classifier training stage on CelebA.} We set $attractiveness$ and $male$ to the target class and sensitive attribute, respectively.}
\label{table:grl2}
\end{table}

\section{Fairness Strategy in Classifier Training Stage}

In Table~\ref{table:grl2}, we explore the effectiveness of applying \textit{GRL}~\cite{grl} in the classifier training stage, after finishing the representation learning with \textit{SupCon} and \textit{FSCL+}. To this end, we deploy an additional classifier for the sensitive attribute and do not freeze the encoder and projection networks in the second stage. As might be expected, \textit{GRL} improves the fairness of \textit{SupCon} by sacrificing the classification accuracy. Meanwhile, it degrades EO as well as the classification accuracy in ours. We speculate that it is because the fair representation learned by \textit{FSCL+} becomes biased by re-training the encoding networks with the cross entropy loss and \textit{GRL}. The similar results of EO and top-1 accuracy between \textit{SupCon} with \textit{GRL} and \textit{FSCL+} with \textit{GRL} support that the learned representation is almost renewed in the classifier training stage. In conclusion, the results show that applying the additional strategy for fairness in the classifier training stage is not effective to our method.

\section{Modification for Incomplete Supervised Setting}
To apply our method to the environment where target class labels are not provided, we introduce \textit{FSCL}\textsuperscript{$\dagger$}, which a modified version of \textit{FSCL}. 
We set a positive sample to another patch from the same image with an anchor and negative samples to $Z_{ig}$ and $Z_{tg}$. It is formulated as follows.
\begin{equation}
    FSCL^{\dagger}=-\sum_{z_i\in Z}\log \frac{exp(z_i\cdot z_{p}/\tau)}{\sum_{z_f^{*}\in Z_f^{*}(i)}exp(z_i \cdot z_f^{*}/\tau)},
\end{equation}      
where $Z_f^{*}(i)=\{z_f^{*} \in Z| \hat{s}_f^{*} = \hat{s}_i\}$. Except for the loss function, the overall structure is the same as the original.

\begin{figure}[h]
  \centering
  \includegraphics[clip=true, width=0.47\textwidth]{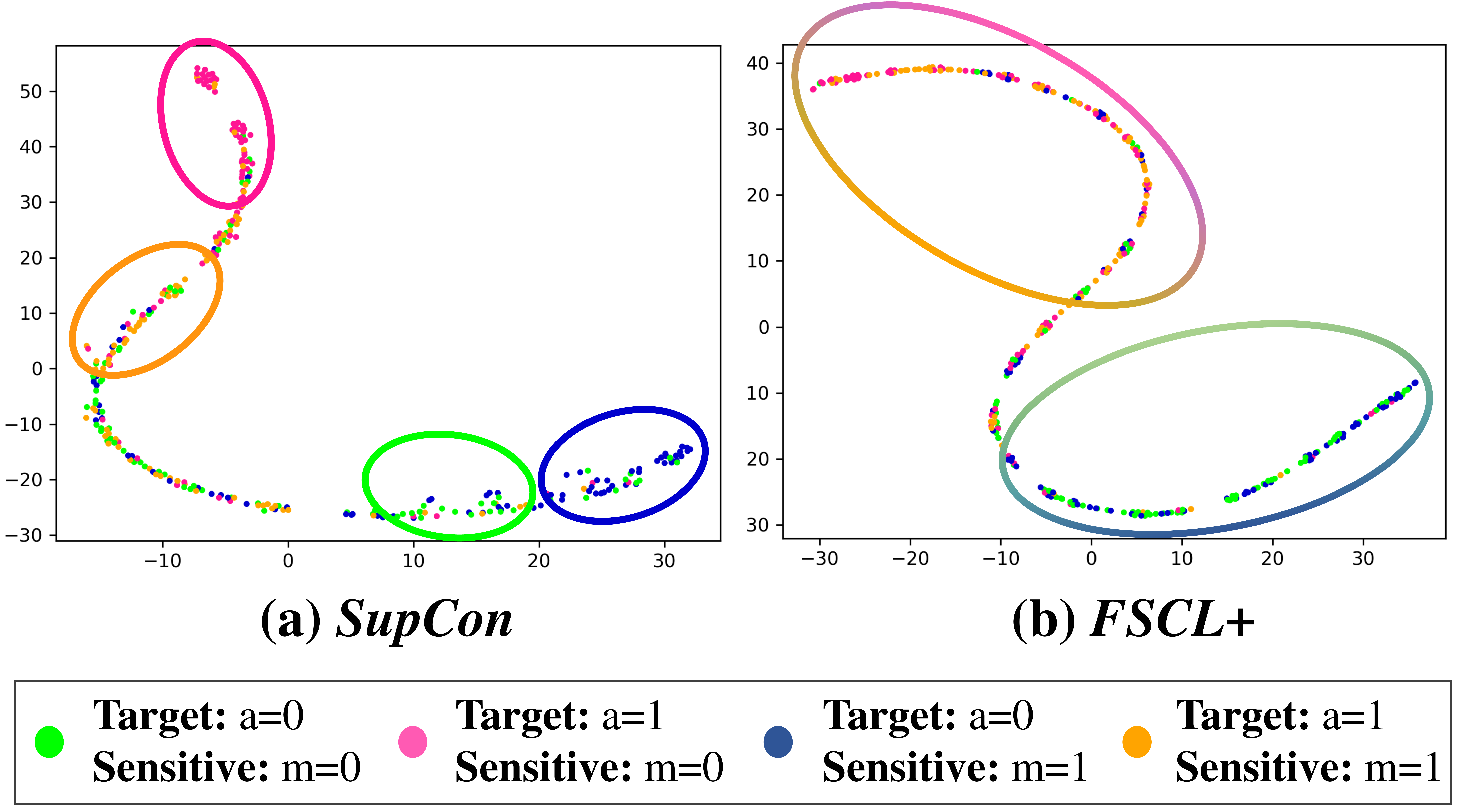}
  \caption{\textbf{t-SNE visualizations with random intialization.}} 
  \label{tsne2}
\end{figure}

\section{Details of t-SNE Visualization}

For the t-SNE~\cite{visualtsne} visualization, we exploit the models pre-trained on CelebA dataset~\cite{celeba} for 100 epochs. Then we obtain 50 random samples (\textit{i.e.}, representation) per data group with the pre-trained models. Before applying the t-SNE algorithm, we reduce the dimensionality of the samples using PCA reduction. We tune the hyperparameters in the scikit-learn implementation as follows.
\begin{itemize}
    \item Perplexity: from 10 to 40 by 1
    \item Learning rate: 10 or 100
    \item Iteration= 100, 1000, or 10000
\end{itemize}    
We set the perplexity, learning rate, and iterations 10, 10, and 10000 respectively, but in all the cases, we note that representation learned by \textit{FSCL+} is more agnostic to the sensitive attribute than that learned by \textit{SupCon}. Furthermore, we provide t-SNE plots without PCA reduction in Figure~\ref{tsne2} since it considerably affects the structure of representations.

\begin{figure}[h]
  \centering
  \includegraphics[clip=true, width=0.47\textwidth]{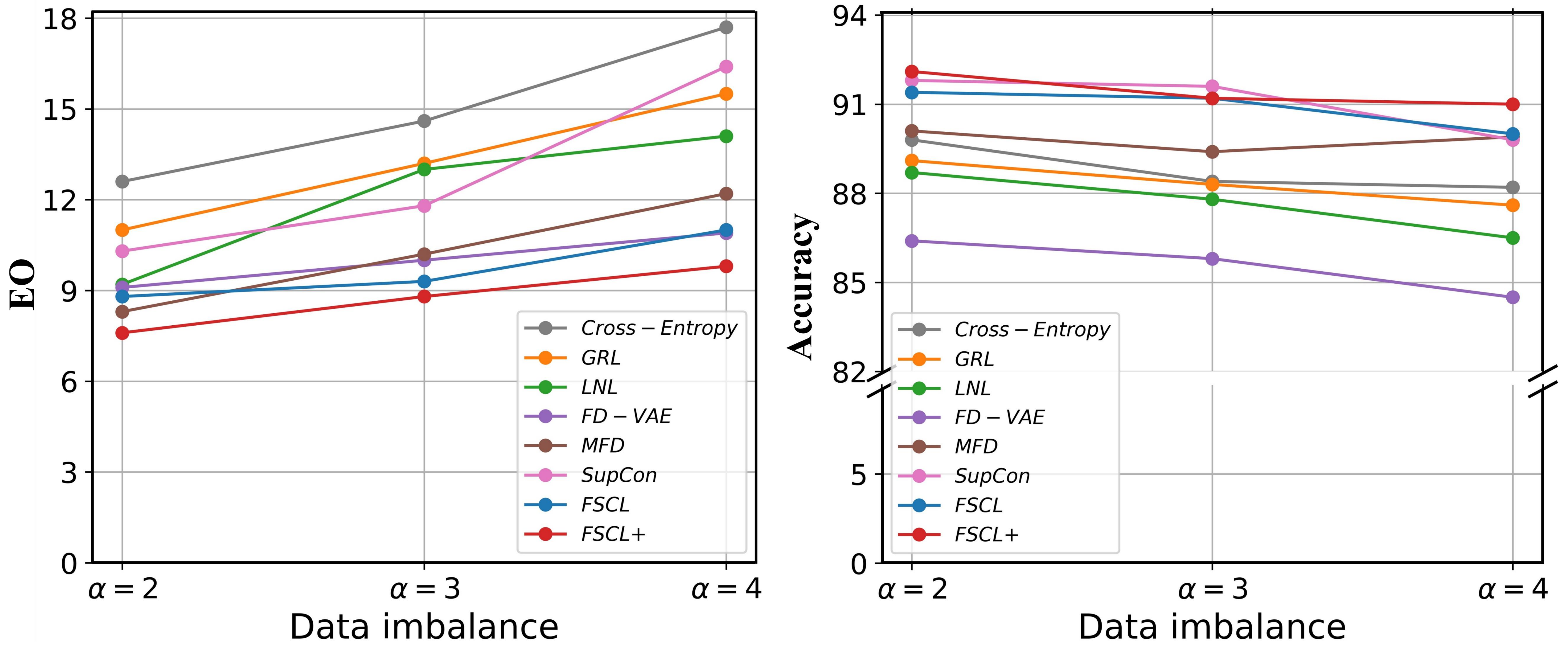}
  \caption{\textbf{Classification results on UTK Face dataset.} We set \textit{gender} and \textit{age} to the target class and sensitive attribute, respectively. It shows trends of classification accuracy and equalized odds (EO) at different $\alpha$.}
  \label{utkface2}
\end{figure}

\begin{figure*}[t]
  \centering
  \includegraphics[clip=true, width=1.0\textwidth]{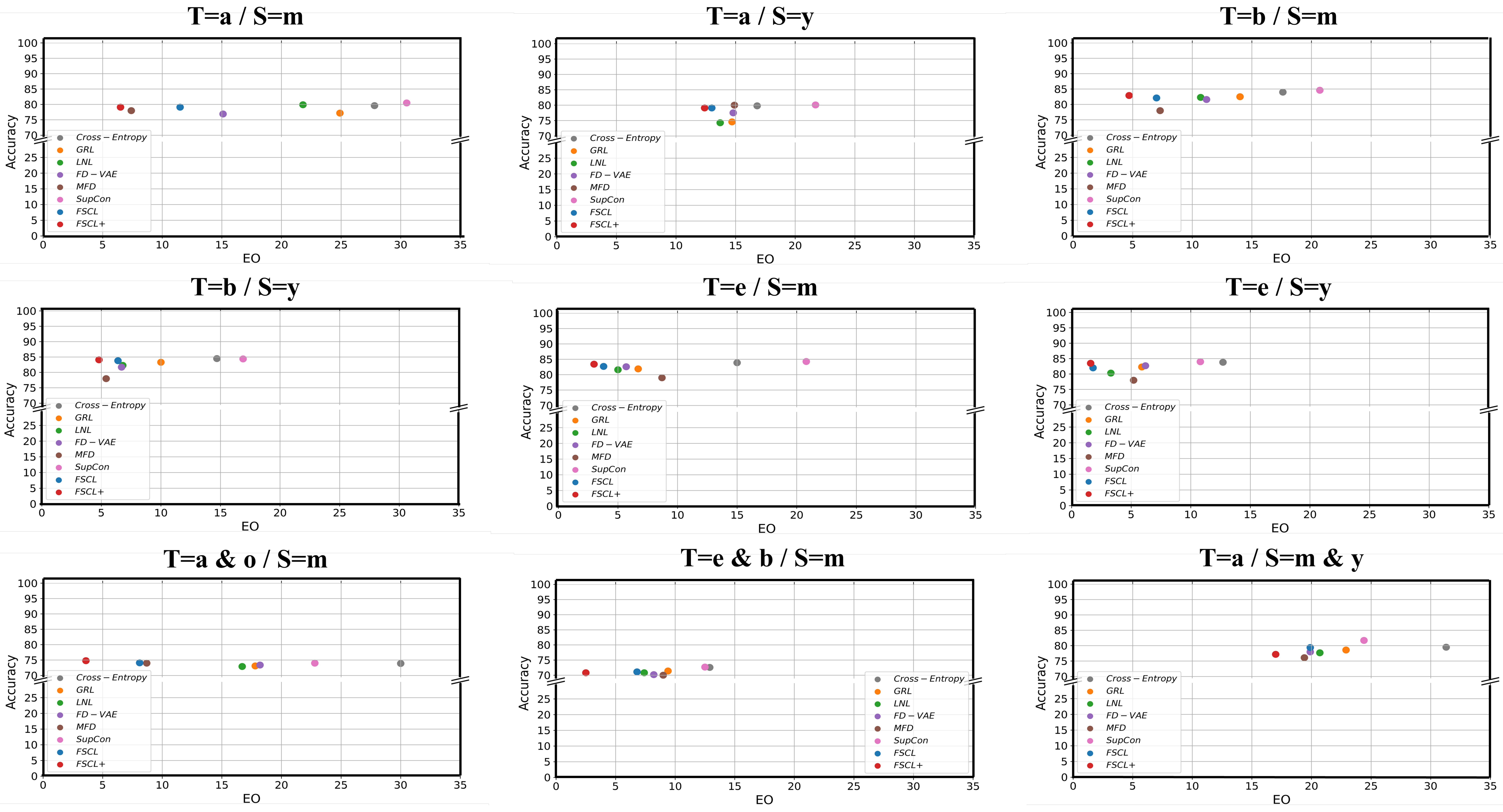}
  \caption{\textbf{Experimental results in figure form on CelebA dataset.} It shows the trade-off performances between ACC. and EO more clearly. The upper left corner of the plots corresponds to the optimal trade-off performance.} 
  \label{celeba}
\end{figure*}
\begin{table*}[t]
\centering
\resizebox{0.99\textwidth}{!}{
\begin{tabular}{cccccccccccccccccccccccc}
\toprule
\multirow{2}{*}{Attributes } & \multicolumn{2}{c}{CE~\cite{resnet}} && \multicolumn{2}{c}{\textit{GRL}~\cite{grl}} && \multicolumn{2}{c}{\textit{LNL}~\cite{lnl}} && \multicolumn{2}{c}{\textit{FD-VAE}~\cite{FDVAE}} && \multicolumn{2}{c}{\textit{MFD}~\cite{MFD}} && \multicolumn{2}{c}{\textit{SupCon}~\cite{supcon}} && \multicolumn{2}{c}{\textit{FSCL} } && \multicolumn{2}{c}{\textit{FSCL}+}  \\ \cmidrule[0.5pt]{2-3} \cmidrule[0.5pt]{5-6} \cmidrule[0.5pt]{8-9} \cmidrule[0.5pt]{11-12} \cmidrule[0.5pt]{14-15} \cmidrule[0.5pt]{17-18} \cmidrule[0.5pt]{20-21} \cmidrule[0.5pt]{23-24}
& EO  & Acc. &&  EO   &  Acc.  && EO    &  Acc.  &&  EO    &  Acc.   && EO    &  Acc.   &&  EO  &  Acc.   &&  EO   &  Acc.   &&  EO   &  Acc.   \\ \cmidrule[0.5pt]{1-24} \morecmidrules\cmidrule[0.5pt]{1-24}
T=\textit{a} / S=\textit{m} & 27.8$_{\pm 0.2}$ & 79.6$_{\pm 0.5}$ && 24.9$_{\pm 0.3}$ &77.2$_{\pm 0.5}$ && 21.8$_{\pm 0.4}$ & 79.9$_{\pm 0.5}$ &&  15.1$_{\pm 0.1}$  & 76.9$_{\pm 0.0}$ && 7.4$_{\pm 0.3}$ & 78.0$_{\pm 0.3}$ && 30.5$_{\pm 1.3}$ &  80.5$_{\pm 0.7}$ &&  11.5$_{\pm 0.3}$  &  79.1$_{\pm 0.4}$  && \textbf{6.5}$_{\pm 0.4}$ & 79.1$_{\pm 0.4}$  \\ \cmidrule[0.5pt]{1-24}
T=\textit{a} / S=\textit{y} & 16.8$_{\pm 0.3}$ & 79.8$_{\pm 0.4}$ && 14.7$_{\pm 0.4}$ & 74.6$_{\pm 0.4}$ && 13.7$_{\pm 0.3}$ & 74.3$_{\pm 0.4}$ && 14.8$_{\pm 0.2}$ & 77.5$_{\pm 0.1}$ && 14.9$_{\pm 0.4}$  & 80.0$_{\pm 0.3}$ && 21.7$_{\pm 1.0}$ & 80.1$_{\pm 0.8}$ && 13.0$_{\pm 0.6}$   & 79.1$_{\pm 0.5}$  && \textbf{12.4}$_{\pm 0.5}$ & 79.1$_{\pm 0.5}$  \\ \cmidrule[0.5pt]{1-24}
T=\textit{b} / S=\textit{m} & 17.6$_{\pm 0.3}$ & 84.0$_{\pm 0.3}$ && 14.0$_{\pm 0.3}$ & 82.5$_{\pm 0.5}$ && 10.7$_{\pm 0.2}$ & 82.3$_{\pm 0.4}$ && 11.2$_{\pm 0.1}$ & 81.6$_{\pm 0.3}$ && 7.3$_{\pm 0.2}$ & 78.0$_{\pm 0.3}$ && 20.7$_{\pm 0.5}$ &84.6$_{\pm 0.6}$ &&  7.0$_{\pm 0.4}$ & 82.1$_{\pm 0.3}$ && \textbf{4.7}$_{\pm 0.5}$ & 82.9$_{\pm 0.4}$   \\ \cmidrule[0.5pt]{1-24}
T=\textit{b} / S=\textit{y} & 14.7$_{\pm 0.1}$ & 84.5$_{\pm 0.3}$ && 10.0$_{\pm 0.2}$ & 83.3$_{\pm 0.5}$ && 6.8$_{\pm 0.3}$ & 82.3$_{\pm 0.5}$ && 6.7$_{\pm 0.2}$ & 81.7$_{\pm 0.0}$ &&5.4$_{\pm 0.1}$ & 78.0$_{\pm 0.2}$ && 16.9$_{\pm 0.9}$ & 84.4$_{\pm 0.8}$ && 6.4$_{\pm 0.4}$ & 83.8$_{\pm 0.4}$ && \textbf{4.8}$_{\pm 0.3}$ & 84.1$_{\pm 0.5}$    \\ \cmidrule[0.5pt]{1-24}
T=\textit{e} / S=\textit{m} & 15.0$_{\pm 0.3}$ & 83.9$_{\pm 0.2}$ && 6.7$_{\pm 0.4}$  & 81.9$_{\pm 0.6}$ && 5.0$_{\pm 0.3}$ & 81.6$_{\pm 0.3}$ && 5.7$_{\pm 0.0}$ & 82.6$_{\pm 0.1}$ && 8.7$_{\pm 0.3}$ & 79.0$_{\pm 0.4}$ && 20.8$_{\pm 1.1}$ & 84.3$_{\pm 0.5}$ && 3.8$_{\pm 0.3}$ & 82.7$_{\pm 0.3}$ && \textbf{3.0}$_{\pm 0.4}$ & 83.4$_{\pm 0.6}$    \\ \cmidrule[0.5pt]{1-24}
T=\textit{e} / S=\textit{y} & 12.7$_{\pm 0.2}$ & 83.8$_{\pm 0.3}$ && 5.9$_{\pm 0.4}$ & 82.3$_{\pm 0.4}$ && 3.3$_{\pm 0.4}$ & 80.3$_{\pm 0.6}$ && 6.2$_{\pm 0.1}$ & 84.0$_{\pm 0.2}$ && 5.2$_{\pm 0.2}$ & 78.0$_{\pm 0.2}$ && 10.8$_{\pm 1.0}$ & 84.0$_{\pm 0.7}$ &&1.8$_{\pm 0.3}$  & 82.0$_{\pm 0.4}$ && \textbf{1.6}$_{\pm 0.3}$ & 83.5$_{\pm 0.3}$   \\ \cmidrule[0.5pt]{1-24}
T=\textit{a \& o} / S=\textit{m} & 30.0$_{\pm 0.2}$ & 73.9$_{\pm 0.5}$ && 17.8$_{\pm 0.2}$  & 73.1$_{\pm 0.5}$ && 16.7$_{\pm 0.4}$ & 72.9$_{\pm 0.5}$ && 18.2$_{\pm 0.1}$ & 73.4$_{\pm 0.1}$ && 8.7$_{\pm 0.4}$ & 74.0$_{\pm 0.3}$ && 22.8$_{\pm 0.7}$  & 74.0$_{\pm 0.5}$ && 8.1$_{\pm 0.3}$ & 74.1$_{\pm 0.3}$ && \textbf{3.6}$_{\pm 0.3}$ & 74.8$_{\pm 0.4}$  \\ \cmidrule[0.5pt]{1-24}
T=\textit{b \& e} / S=\textit{m} & 12.9$_{\pm 0.2}$ & 72.6$_{\pm 0.4}$ && 9.4$_{\pm 0.3}$ & 71.4$_{\pm 0.4}$ && 7.4$_{\pm 0.2}$ & 70.8$_{\pm 0.5}$ && 8.2$_{\pm 0.1}$ & 70.2$_{\pm 0.2}$ && 9.0$_{\pm 0.1}$ & 70.0$_{\pm 0.1}$ && 12.5$_{\pm 0.8}$ & 72.7$_{\pm 0.9}$ && 6.8$_{\pm 0.4}$ & 71.1 $_{\pm 0.2}$&& \textbf{2.5}$_{\pm 0.6}$ & 70.8$_{\pm 0.5}$   \\ \cmidrule[0.5pt]{1-24}
T=\textit{a} / S=\textit{m \& y} & 31.3$_{\pm 0.3}$ & 79.5$_{\pm 0.4}$ && 22.9$_{\pm 0.4}$ & 78.6$_{\pm 0.5}$ && 20.7$_{\pm 0.3}$ & 77.7$_{\pm 0.5}$ && 19.9$_{\pm 0.0}$ & 78.0$_{\pm 0.1}$ && 19.4$_{\pm 0.2}$ & 76.1$_{\pm 0.3}$ && 24.4$_{\pm 1.3}$  & 81.7$_{\pm 0.7}$ && 19.9$_{\pm 0.5}$ & 79.4$_{\pm 0.3}$&& \textbf{17.0}$_{\pm 0.5}$ & 77.2$_{\pm 0.5}$  \\ \bottomrule
\end{tabular}
}
\caption{\textbf{Classification results on CelebA.} We further specify the standard deviation in this table.}
\label{table:main2}
\end{table*}

\section{Further Experiments on UTK Face}

In Figure \ref{utkface2}, we provide experimental results on UTK Face with the other sensitive attribute, \textit{age}. It shows that \textit{FSCL+} maintain the fairest EO and the best top-1 accuracy at all $\alpha$. Although FD-VAE~\cite{FDVAE} achieves similar EO with \textit{FSCL}, its accuracy is significantly inferior to ours. It indicates that ours highly outperform it in terms of the trade-off performance between fairness and accuracy.

\section{Additional Experimental Results on CelebA}
To clearly show the trade-off performances between classification accuracy and fairness, we plot the experimental results on CelebA in Figure~\ref{celeba}. \textit{FSCL+} achieves the best trade-off performances in all the results. Furthermore, we supplement the experimental results by reporting standard deviation in Table~\ref{table:main2}.

\begin{table*}[h]
\centering
\resizebox{0.8\textwidth}{!}{
\begin{tabular}{ccccccccc} \toprule
\multicolumn{9}{c}{CelebA} \\ \cmidrule[0.5pt]{1-9}\morecmidrules\cmidrule[0.5pt]{1-9}
 & \textit{a}=0 & \textit{a}=1 &  & \textit{b}=0 & \textit{b}=1 &  & \textit{e}=0 & \textit{e}=1 \\ \hline
\textit{m}=0 & 29,920 & 64,589 & \textit{m}=0 & 84,954 & 9,555 & \textit{m}=0 & 84,963 & 9,546 \\
\textit{m}=1 & 49,247 & 19,014 & \textit{m}=1 & 39,475 & 28,786 & \textit{m}=1 & 44,527 & 23,734 \\ \cmidrule[0.5pt]{1-9}\morecmidrules\cmidrule[0.5pt]{1-9}
 & \textit{a}=0 & \textit{a}=1 &  & \textit{b}=0 & \textit{b}=1 &  & \textit{e}=0 & \textit{e}=1 \\ \hline
\textit{y}=0 & 30,618 & 5,364 & \textit{m}=0 & 19,164 & 16,818 & \textit{m}=0 & 22,146 & 13,836 \\
\textit{y}=1 & 48,549 & 78,239 & \textit{m}=1 & 105,265 & 21,523 & \textit{m}=1 & 107,344 & 19,444 \\ \cmidrule[0.5pt]{1-9}\morecmidrules\cmidrule[0.5pt]{1-9}
 & \textit{a}=0 & \textit{a}=1 &  & \textit{m}=0 & \textit{m}=1 &  & \textit{m}=0 & \textit{m}=1 \\ \hline
\textit{m}=0, \textit{y}=0 & 7,522 & 3,645 & \textit{a}=0, \textit{o}=0 & 13,995 & 27,966 & \textit{b}=0, \textit{e}=0 & 78,613 & 30,481 \\
\textit{m}=1, \textit{y}=0 & 23,096 & 1,719 & \textit{a}=1, \textit{o}=0 & 30,943 & 11,380 & \textit{b}=1, \textit{e}=0 & 6,350 & 14,046 \\
\textit{m}=0, \textit{y}=1 & 22,398 & 60,944 & \textit{a}=0, \textit{o}=1 & 15,925 & 21,281 & \textit{b}=0, \textit{e}=1 & 6,341 & 8,994 \\
\textit{m}=1, \textit{y}=1 & 26,151 & 17,295 & \textit{a}=1, \textit{o}=1 & 33,646 & 7,634 & \textit{b}=1, \textit{e}=1 & 3,205 & 14,740\\
\bottomrule
\end{tabular}
}

\caption{\textbf{Composition of the training set of CelebA.} \textit{a}, \textit{b}, \textit{e}, \textit{o}, \textit{m}, and \textit{y} denote \textit{attractiveness}, \textit{bignose}, \textit{bags-under-eyes}, \textit{mouth-slightly-open}, \textit{male}, and \textit{young}, respectively.}
\label{table:celeba}
\end{table*}

\begin{table*}[h]
\centering
\resizebox{0.8\textwidth}{!}{
\begin{tabular}{cccccc}
\toprule
\multicolumn{6}{c}{UTK Face} \\ \cmidrule[0.5pt]{1-6}\morecmidrules\cmidrule[0.5pt]{1-6}
\multicolumn{6}{c}{$\alpha=2$ / $\alpha=3$ / $\alpha=4$} \\ \hline
 & \multicolumn{2}{c}{Ethinicity} &  & \multicolumn{2}{c}{Age} \\ \cline{2-3} \cline{5-6}
 & Caucasian & Others &  & More than 35 & Others \\ \cline{1-3} \cline{5-6} 
Female & 1,666 / 1,250 / 1,000 & 3,334 / 3,750 / 4,000 &  & 1,666 / 1,250 / 1,000 & 3,334 / 3,750 / 4,000 \\
Male & 3,334 / 3,750 / 4,000 & 1,666 / 1,250 / 1,000 &  & 3,334 / 3,750 / 4,000 & 1,666 / 1,250 / 1,000 \\
\bottomrule
\end{tabular}
}

\caption{\textbf{Composition of the training set of UTK Face.} $\alpha$ denotes the intensities of data imbalance.}
\label{table:utk}
\end{table*}

\section{Dataset Composition}

\subsection{CelebA and UTK Face}
In CelebA~\cite{celeba}, we conduct experiments in terms of a variety of target and sensitive attribute pairs. Table~\ref{table:celeba} shows the specific composition of the training set in all the settings. In UTK Face~\cite{utkface}, we involve 10,000, 2,400, and 2,400 data in the training, validation, and test sets, respectively. We provide the various compositions of the training set according to $\alpha$ in Table~\ref{table:utk}.

\subsection{Dogs and Cats}
Similar to UTK Face, we leverage 3,425 black cat and white dog images, and 685 white cat and black dog images for training. The test set includes 2,400 images which are completely balanced. We note that it is different from the original setting in~\cite{lnl}. In the study, the target attribute and bias are completely correlated in the training set. For instance, cats are always black and dogs are always white. Although they solved the task by utilizing the pixel-level of bias labels (\textit{i.e.}, RGB values of each pixel), it is an almost unsolvable problem with only the image-level of labels since the target attribute and bias labels are always the same at the image-level. Therefore, we designed the task more reasonable to validate fairness methods which mostly exploit the image-level of labels.

\subsection{Discussion on License and Data Collection}
Both CelebA~\cite{celeba} and UTK Face~\cite{utkface} have a non-standard license (i.e, Custom (non-commericial)), but the creators clarify the datasets are available for non-commercial research purposes only.

CelebA consists of the images collected from Celeb-Faces dataset~\cite{celebfaces} and attribute labels. According to \cite{celebfaces}, the images are collected by searching names of celebrities on the web. Also in UTK Face, the creators combine the images from CACD~\cite{cacd} and Morph~\cite{morph} datasets with the images crawled in Bing and Google search engines. In both CACD and Morph, the images are gathered by searching on the web. 


\section{Implementation Details}
\subsection{Structure of Comparable Models}
\textbf{\textit{Cross-Entropy}}~\cite{resnet} , \textbf{\textit{GRL}}~\cite{grl}, \textbf{\textit{LNL}}~\cite{lnl}: The models utilize ResNet-18~\cite{resnet} for backbone networks and a MLP with one hidden layer for classifiers. The dimensions of representation are the same as ours. \textit{GRL} and \textit{LNL} are reproduced based on~\cite{grl,lnl}, and the hyperparameter to determine a weight for the reversed gradient is searched in the range from 0.01 to 0.1 in each experiment. For \textit{LNL}, hyperparameter $\lambda$ for regularization loss is searched in the range from 0.01 to 0.1 in each experiment. For all the models, we train them in an end-to-end manner for 100 epochs.

\textbf{\textit{FD-VAE}}~\cite{FDVAE}: We build the model with the same structure as the original paper~\cite{FDVAE} without the encoder network. For a fair comparison, we substitute the encoder network to ResNet-18 and obtain better reproduction performances. Following the paper, we separate each latent space to have the same dimensions to each other and set hyperparameter $\beta$ to 1. The other hyperparameters are found by grid searching and set to $\alpha=1$, $\gamma=5$, and $\lambda=1$ for all the experiments. For representation learning, we train the encoder networks for 100 epochs. After that, we train the classifiers for downstream tasks for 10 epochs.

\textbf{\textit{MFD}}~\cite{MFD}: We implement the model with source code released by the authors. The teacher and student models both leverage ResNet-18 for backbone networks and a MLP with one hidden layer for a classifier. Following the original paper, we train the models for 50 epochs and set hyperparameter $\lambda$ to 7 and 5 for CelebA and UTK Face, respectively. For Dogs and Cats, $\lambda$ is determined as 7 through grid searching. 

\textbf{\textit{SupCon}}~\cite{supcon}, \textbf{\textit{SimCLR}}~\cite{simclr}, \textbf{\textit{FSCL} (ours)}: We implement \textit{SupCon} and \textit{SimCLR} with source code released by the authors of~\cite{supcon}, and \textit{FSCL} is also based on the code (which is licensed under the terms of the MIT license). The models use ResNet-18~\cite{resnet} for the encoder network and a MLP with two hidden layers for the projection network, which have 256 hidden nodes.

\subsection{Augmentation Strategy and Experimental Setup}
 
For the models based on contrastive loss, we augment two patches per image. Except for this, we use the same augmentation strategy~\cite{simclr} for all the models. Specifically, we sequentially and randomly apply cropping and resizing, horizontal flipping, color jittering, and gray scaling. 

For all the models, we set the identical environments of SGD optimizer with momentum~\cite{optimizer}, batch sizes of 128, and learning rate of 0.1. All the experiments are based on the PyTorch library and are conducted in a Linux environment with 4 NVIDIA Titan Xp GPUs with 12GB of memory.

\begin{table}[h]
\centering
\resizebox{0.45\textwidth}{!}{
\begin{tabular}{cccc} \toprule
Method & Regularization & EO ($\downarrow$) & Acc. ($\uparrow$) \\ \cmidrule[0.5pt]{1-4} \morecmidrules\cmidrule[0.5pt]{1-4}
\multirow{5}{*}{\textit{GDRO}} 
 & Standard & 21.3$_{\pm 1.0}$ & 76.3$_{\pm 0.2}$ \\ \morecmidrules\cmidrule[0.3pt]{2-4}
 & Early Stopping & \textbf{4.0} $_{\pm 0.1}$ & 74.7$_{\pm 0.1}$ \\ \morecmidrules\cmidrule[0.3pt]{2-4}
 & Strong $L_2$ (lr=0.1) & 8.7 $_{\pm 2.6}$ & 76.3$_{\pm 0.1}$ \\ \morecmidrules\cmidrule[0.3pt]{2-4}
  &Strong $L_2$ \&  Group adjustments (C=5) & 8.0 $_{\pm 2.0}$& 77.1 $_{\pm 0.2}$\\ \morecmidrules\cmidrule[0.5pt]{1-4}
 \textit{FSCL+} & Standard & 6.5$_{\pm 0.4}$ & 79.1$_{\pm 0.1}$ \\ \bottomrule
\end{tabular}

}
\caption{\textbf{Comparison with GDRO on CelebA.} We set \textit{attractiveness} and \textit{male} to the target class and sensitive attribute, respectively.}
\label{table:gdro}
\end{table}

\section{Comparison with GDRO}
GDRO~\cite{GDRO} is one of the state-of-the-art methods to minimize the performance gaps between data groups and has a goal similar to our group-wise normalization. Thus, we report comparison results with GDRO in Table~\ref{table:gdro}. Following the original paper, we search for the best C in the range of [0, 5]. The results show that ours achieves a better trade-off performance than GDRO.



\section{Two kinds of Supervised Contrastive Losses}

In this section, we summarize two kinds of supervised contrastive losses (\textit{i.e.}, $L^{sup}_{out}$ and $L^{sup}_{in}$) proposed in ~\cite{supcon} and why we leverage $L^{sup}_{out}$ as our baseline. Unlike $L^{sup}_{out}$ (\textit{i.e.}, $L^{Sup}$ in the main paper), $L^{sup}_{in}$ places the summation over positive samples and the normalization factor inside the log as follows.
\begin{equation}\label{supcon}
    L^{Sup}_{in}=-\sum_{z_i\in Z}\log \bigg ( \frac{1}{\left | Z_p(i) \right |} \sum_{z_p\in Z_p(i)}\frac{\phi_{p}}{\sum_{z_a\in Z_a(i)}\phi_{a}} \bigg ).
\end{equation} 
In the loss, the normalization factor works as a constant (\textit{i.e.}, $-\sum_{z_i\in Z}\log \frac{1}{\left | Z_p(i) \right |}$), so it cannot normalize the imbalance in the positive samples. As the result, $L^{sup}_{in}$ is more vulnerable to the data bias and shows inferior classification performances to $L^{sup}_{out}$. For these reasons, we utilize the latter as our baseline.

\section{Discussion on Limitations}

In this section, we discuss two limitations of our study. The first one is that our work is confined to the image classification task. We discuss it by explaining why we cover the task in this paper. One reason is that the superior performance of our baselines (\textit{i.e}., SupCon and SimCLR) has been experimentally validated in the image classification task~\cite{supcon,simclr}. Therefore, through the task, we can make a fair comparison with the models and convincingly demonstrate our improvement over them. The other reason is that image classification is a fundamental and common task not only in contrastive representation learning~\cite{simclr,supcon,multiview,memorybank} but in fairness studies in the field of computer vision~\cite{FFVAE,FDVAE,towardfairness,fairnessgan}. Although fair visual representation can be exploited in other tasks, such as object recognition~\cite{balance}, image-to-image translation~\cite{bmvc_ours,icassp}, face recognition~\cite{gendershades,debface}, and object detection~\cite{fairdetection}, each of them requires a suitable notion of fairness~\cite{balance,fairdetection} and specialized architectures~\cite{debface,icassp,bmvc_ours}. Therefore, to achieve the best performance on the tasks, we also need to modify the proposed loss more appropriately for them.  We leave the extension of~\textit{FSCL} to broader tasks for future work.

Second, our method essentially requires sensitive attribute labels to improve fairness. Even though supervision of the sensitive attribute labels is common in the literature on fair classification~\cite{AdversarialFair,FFVAE,FDVAE,fairnessgan}, sometimes we cannot access the labels and it is laborious and expensive to annotate them. Although we show that our method can reduce such costs by effectively improving fairness using only a few labels, it cannot be utilized in the complete absence of the labels. Therefore, future works that develop a fair contrastive loss free of the sensitive attribute labels would make a significant contribution to the research community. We expect our study to be a bridgehead for them.

\end{document}